\documentclass[lettersize,journal]{IEEEtran}
\usepackage{amsmath}
\usepackage{amsfonts}
\usepackage{algorithmic}
\usepackage{algorithm}
\usepackage{array}
\usepackage{graphicx}
\usepackage{subfigure} 
\usepackage{textcomp}
\usepackage{stfloats}
\usepackage{url}
\usepackage{verbatim}
\usepackage{cite}
\usepackage[percent]{overpic}
\usepackage{booktabs}
\usepackage{multirow}
\usepackage{threeparttable}
\usepackage[colorlinks,
            linkcolor=red,
            anchorcolor=blue,
            urlcolor=magenta,
            citecolor=blue]{hyperref}
\usepackage[font=footnotesize]{caption} 
\usepackage{colortbl}
\usepackage{ulem}

\usepackage{xcolor}

\newcommand{\CSRED}[1]{\textcolor[RGB]{171, 36, 45}{#1}}
\newcommand{\CSBLUE}[1]{\textcolor[RGB]{45, 77, 152}{#1}}
\definecolor{rblue}{rgb}{0,0,0}
\newcommand{\add}[1]{{\textcolor{rblue}{#1}}}
\newcommand{\changed}[1]{{\textcolor{rblue}{#1}}}
\definecolor{rrblue}{rgb}{0,0,0}
\newcommand{\addROne}[1]{{\textcolor{rrblue}{#1}}}
\newcommand{\changedROne}[1]{{\textcolor{rrblue}{#1}}}

\begin{document}

\title{P2U-SLAM:\\A Monocular Wide-FoV SLAM System Based on Point Uncertainty and Pose Uncertainty}

\author{Yufan Zhang$^{1}$, Kailun Yang$^{2,3}$, Ze Wang$^{1}$, and Kaiwei Wang$^{1,*}$
\thanks{This was supported in part by the National Natural Science Foundation of China (Grant No. 12174341 and No. 62473139), in part by the Hunan Provincial Research and Development Project (Grant No. 2025QK3019), and in part by the Open Research Project of the State Key Laboratory of Industrial Control Technology, China (Grant No. ICT2025B20), in part by the State Key Laboratory of Autonomous Intelligent Unmanned Systems (the opening project number ZZKF2025-2-10), in part by Hangzhou SurImage Technology Company Ltd., and in part by Hangzhou HuanJun Technology Company Ltd.}
\thanks{Y. Zhang, Z. Wang, and K. Wang are with the State Key Laboratory of Extreme Photonics and Instrumentation, Zhejiang University, Hangzhou 310027, China.}
\thanks{$^{2}$K. Yang is with the School of Artificial Intelligence and Robotics and the National Engineering Research Center of Robot Visual Perception and Control Technology, Hunan University, Changsha 410012, China.}%
\thanks{$^{3}$K. Yang is also with the National Engineering Research Center of Robot Visual Perception and Control Technology, Hunan University, Changsha 410082, China.}%
\thanks{$^*$Corresponding author: Kaiwei Wang (E-mail: wangkaiwei@zju.edu.cn).}
}

\markboth{IEEE Transactions on Intelligent Transportation Systems, January~2026}%
{Zhang \MakeLowercase{\textit{et al.}}: P2U-SLAM}

\maketitle

\begin{abstract}
This paper presents P2U-SLAM, a visual Simultaneous Localization And Mapping (SLAM) system with a wide Field of View (FoV) camera, which utilizes pose uncertainty and point uncertainty. While the wide FoV enables considerable repetitive observations of historical map points for matching cross-view features, the data properties of the historical map points and the poses of historical keyframes have changed during the optimization process. The neglect of data property changes results in the lack of partial information matrices in optimization, increasing the risk of long-term positioning performance degradation. The purpose of our research is to mitigate the risks posed by wide-FoV visual input to the SLAM system. Based on the conditional probability model, this work reveals the definite impacts of the above data properties changes on the optimization process, concretizes these impacts as point uncertainty and pose uncertainty, and gives their specific mathematical form. P2U-SLAM embeds point uncertainty into the tracking module and pose uncertainty into the local mapping module respectively, and updates these uncertainties after each optimization operation including local mapping, map merging, and loop closing. We present an exhaustive evaluation on $27$ sequences from two popular public datasets with wide-FoV visual input. P2U-SLAM shows excellent performance compared with other state-of-the-art methods. The source code will be made publicly available at \url{https://github.com/BambValley/P2U-SLAM}.

\end{abstract}


\section{Introduction}
\begin{figure}[t!]
\centering
\includegraphics[width=1\linewidth]{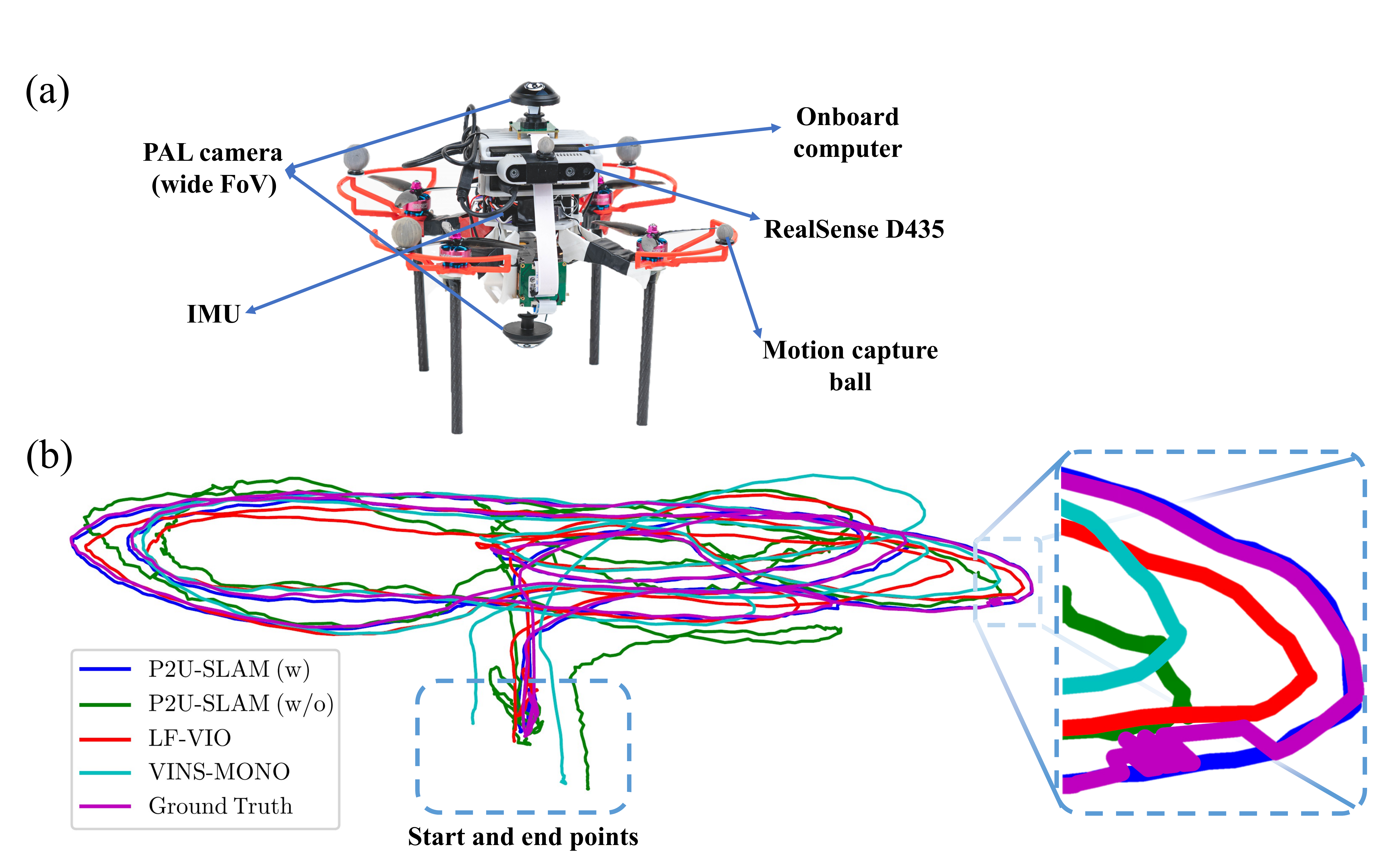}
\caption{An example of the wide-FoV lens used in transportation systems. a) The aerial platform used for data collection in the PALVIO dataset~\cite{wang2022lf}, equipped with two wide-FoV Panoramic Annular Lens (PAL) cameras; b) The output trajectories of several SLAM/VIO algorithms on the ID01 sequence of the PALVIO dataset, where the blue and green trajectories are the results of the P2U-SLAM proposed in this paper. In addition, (\textit{w.}) indicates running with the support of point uncertainty and pose uncertainty, and (\textit{w./o.}) indicates running without that support.
}
\label{fig:exibition_for_mobile_platform}
\end{figure}

\IEEEPARstart{I}{ntensive} investigation on Visual Simultaneous Localization and
Mapping (SLAM), a method utilizing sensors to estimate agent pose and establish a map of environment~\cite{cadena2016past,forster2014svo,forster2016svo,engel2017direct,2017Stereo,zubizarreta2020direct,mur2015orb,mur2017orb,qin2018vins,10461980,9994237,10591477,10542573,10260268,Ge2024PIPOSLAMLV,Xu2022D2SLAMDA,Boche2025OKVIS2XOK}, is emerging for decades, in the field of autonomous navigation and intelligent transportation system.
Given the robustness of many research works on SLAM suffering from rapid motion and dynamic scenarios, wide Field of View (FoV) cameras (Fig.~\ref{fig:exibition_for_mobile_platform}), such as panoramic annular cameras or fisheye cameras, have become an important type of sensor for SLAM application in light of their natural wider overlap area across frames~\cite{7009773,9554752,8813985,10610351,hu2019indoor,chen2019palvo,chen2021panoramic,wang2022pal}.
 
There have been multiple visual SLAM (vSLAM) frameworks~\cite{campos2021orb,hu2019indoor,matsuki2018omnidirectional,chen2019palvo,chen2021panoramic,wang2022pal, wang2024pal,wang2022lf} owning the ability to apply wide-FoV cameras and obtain progress of robustness for enhancing pose estimation in real-world transportation scenarios. 
These frameworks have realized the adaptation of the nonlinear wide-FoV camera models and the corresponding epipolar constraint due to nonlinear models.

However, the impact of the switch from ordinary pinhole cameras to wide-FoV cameras, which involves many aspects of SLAM, should be more comprehensively investigated.
In the process of matching features in the front-end of vSLAM, different FoV distribution of the same feature induces a huge difference of distortion in co-visible frames~\cite{zhao2015sphorb,urban2017mdbrief}.
This difference of distortion results in a change of the neighborhood grayscale arrangement of the 2D feature in a wide-FoV image~\cite{zhu2018robust}, and leads to weak visual constraints for feature-based methods. 
On the other hand, a wider overlap area produces more observations of 3D points between keyframes, from which Bundle Adjustment (BA) in the SLAM system could benefit~\cite{chen2019palvo}. 
At the same time, additional weak constraints might be introduced as the same 3D point could be observed by two keyframes that are much farther apart from each other,
and the uncertainty of 3D points and co-visible keyframe poses will be more crucial when the camera is re-approaching a historical observation point~\cite{dai2020rgb}. 
Though those weak constraints are enhanced or eliminated within several turns of BA optimization, the accuracy and efficiency of pose estimation have been influenced~\cite{park2022keeping}. 
Some frameworks~\cite{2003Real,2007MonoSLAM,2007A,forster2014svo,chen2019palvo} of direct or semi-direct visual SLAM suppress the impact of point uncertainty through a Bayesian framework before these immature points are fed into the process of BA, yet, the Bayesian framework is only effective within a short-term window of a few seconds. It remains powerless against the effects of point re-observation in medium- or long-term tracking, which could last for hours.

In view of this, we propose P2U-SLAM, a feature-based wide-FoV SLAM framework based on point uncertainty and pose uncertainty. 
Unlike the fixed round 2D covariance used for conventional SLAM frameworks such as ORB-SLAM3~\cite{campos2021orb}, PAL-SLAM~\cite{wang2022pal}, and so on, P2U-SLAM utilizes the ellipsoidal 3D point uncertainty, which can be combined with the Jacobian matrix of the camera model to adapt to changes in neighborhood distribution caused by distortion of wide-FoV images. 
In the meantime, point uncertainty and pose uncertainty can acquire historical observation errors to instruct the direction of optimization. 
They are quantified as a covariance matrix and integrated into the information matrix of BA so that weak constraints introduced by weak feature correspondence from a wide overlap area will be suppressed at any time of the BA process.
P2U-SLAM adopts a multi-threading solution to ensure operation efficiency, mainly including tracking, local mapping, and loop detection threads. Point uncertainty and pose uncertainty are applied in the tracking thread and local mapping thread respectively to improve the accuracy and efficiency of the optimization process.

We verify the proposed algorithm on two public datasets with different kinds of wide-FoV cameras. On the public PALVIO dataset with a drone platform, P2U-SLAM achieves a $59.7\%$ reduction in mean Absolute Trajectory Error (ATE) over the second-best method. On the TUM-VI dataset with a mobile platform, P2U-SLAM is shown to maintain comparable cumulative drift over long-term sequences as short-term indoor sequences, which far exceeds other state-of-the-art methods. Moreover, a set of multi-session experiments has been run on P2U-SLAM with three environments from the TUM-VI dataset to show its map reuse capabilities.

The contributions of this article are summarized as follows:
\begin{enumerate}
\item{This article proposes P2U-SLAM, which is, to the best of the authors' knowledge, the first visual wide-FoV SLAM system able to benefit from point and pose uncertainty. To enhance robustness, point- and pose uncertainty are applied to all SLAM optimization tasks: tracking, mapping, relocalization, and loop closing. The projection of 3D point uncertainty is flexible to distortion of wide-FoV images with the help of the nonlinear wide-FoV camera model, such as the Taylor camera model and Kannala-Brandt fisheye model.}
\item{The calculation approach of point/pose uncertainty is proposed to overcome the shortcoming that the point/pose uncertainty introduces weak constraints for a wider overlap area, which is quantified as a 3D/6D covariance matrix, and it can flexibly converse to the 2D version with the Jacobian matrix of the camera model's project function.}
\item{A complete SLAM framework designed for a wide-FoV camera is established based on previous work, in which initialization progress, feature correspondence, and bundle adjustment are all specially tailored or redesigned for wide-FoV cameras.}
\item{Through extensive experiments on public datasets with different wide-FoV inputs, P2U-SLAM has been proven to outperform state-of-the-art SLAM and VIO methods in terms of robustness, positioning accuracy, and long-term tracking capabilities.}
\end{enumerate}

The remainder of this paper is organized as follows. 
Sec.~\ref{sec:related_work} provides a brief review of related works. 
\changed{Sec}.~\ref{sec:camera_model} introduces the wide-FoV camera models P2U-SLAM supports. 
\changed{Sec}.~\ref{sec:problem_statement} explains why point uncertainty and pose uncertainty are important and how to use them in detail. 
A system overview of P2U-SLAM is provided in Sec.~\ref{sec:system_overview}. 
In Sec.~\ref{sec:experiments}, a comprehensive set of experiments is performed to evaluate our algorithm. 
Finally, the conclusions are drawn and future research directions are pointed out in Sec.~\ref{sec:conclusion}.

\section{Related Work}
\label{sec:related_work}

A brief review of advancements is given in Visual and Visual-Inertial Odometry (VO/VIO) and wide-FoV SLAM frameworks in this section.

\subsection{VO/VIO and SLAM}
Visual odometry is the process of computing the six Degree-of-Freedom (DoF) pose by using vision measurement units. 
Engel~\textit{et al.}~\cite{engel2017direct} proposed Direct Sparse Odometry (DSO), a classic VO framework using direct methods, which directly exploits pixel intensity values instead of artificial vision features. 
Forster~\textit{et al.}~\cite{forster2014svo} proposed Semi-direct monocular Visual Odometry (SVO), which realized a combination of direct and feature-based methods.
Visual-inertial odometry is a VO framework that utilizes vision sensors and inertial measurement units.
Qin~\textit{et al.} proposed VINS-Mono~\cite{qin2018vins}, a robust and universal monocular vision inertial state estimator. 
Then, SVO2.0~\cite{forster2016svo}, the visual-inertial version of SVO, has been designed for monocular and multi-camera systems.
SLAM can be regarded as the VO/VIO with loop closure and global optimization after years of development. 
Mur-Artal~\textit{et al.} designed a novel monocular SLAM system named ORB-SLAM~\cite{mur2015orb}. 
Nowadays, ORB-SLAM has been upgraded to ORB-SLAM3~\cite{campos2021orb}, an accurate open-source library for vision, visual-inertial, and multi-map SLAM.
Similarly, VINS-Fusion~\cite{qin2018online,qin2019vins} has realized the ability of loop closure to become a complete SLAM system based on VINS-Mono.

Nevertheless, most of the above frameworks are not able to work well with wide-FoV camera systems featuring $360^{\circ}$ omnidirectional FoV in actual deployment in autonomous navigation systems. 
Therefore, there are plenty of recent works aiming at the adaptation of nonlinear wide-FoV cameras, which even have a negative half plane with FoV beyond $180${\textdegree}.

\begin{figure}[t!]
\centering
\includegraphics[width=3.4in]{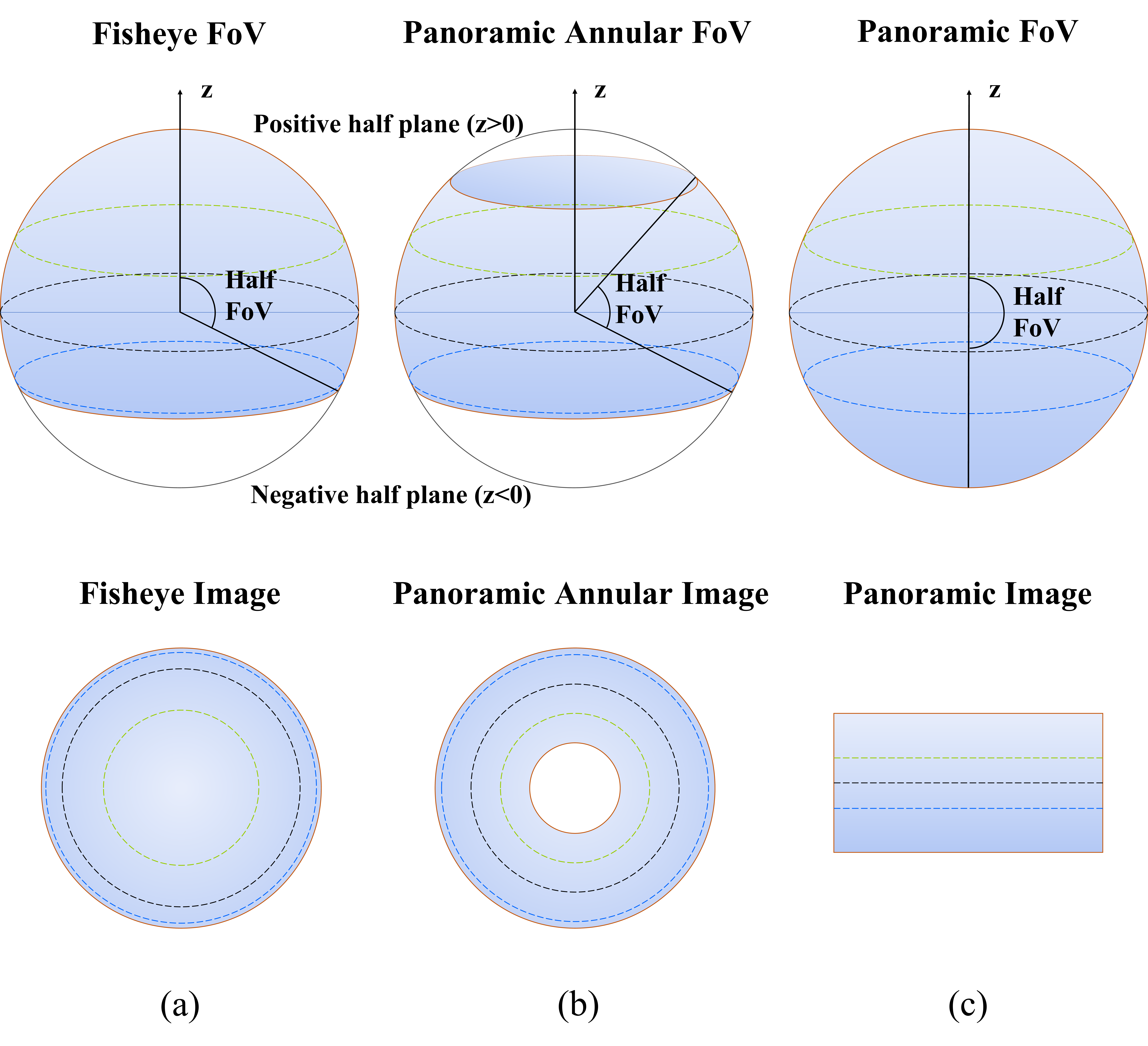}
\caption{Several kinds of wide-FoV images and their FoV distribution. a) The radial FoV distribution of the fisheye image is generally nonlinear; b) the reflective structure of PAL leads to the presence of a central blind area in PAL images, while it gains a larger optical design space to ensure a nearly linear radial FoV distribution; c) the panoramic image is a rectangle and its complete FoV is usually projected and stitched together with multiple cameras.}
\label{fig_wide_fov_cam}
\end{figure}

\subsection{Wide-FoV SLAM}
The wide-FoV SLAM exhibits high stability of localization in situations such as rapid motion and scene transitions. Common wide-FoV lenses mainly include fisheye lenses~\cite{9913937}, catadioptric cameras~\cite{1460954}, multi-camera panoramic imaging systems, and Panoramic Annular Lens (PAL) cameras~\cite{9927463,Wang2022HighperformancePA}. 
The Omnidirectional DSO~\cite{matsuki2018omnidirectional} was designed by Matsuki~\textit{et al.}, and it combines the MEI camera model~\cite{geyer2000unifying} and DSO to build a direct VO that supports the omnidirectional camera.
ORB-SLAM3~\cite{campos2021orb} utilized the Kannala-Brandt fisheye model~\cite{kannala2006generic} based on their previous work to support the fisheye lens. 
Panoramic Visual Odometry (PVO)~\cite{8610700} was proposed for the Ricoh Theta V panoramic camera with a panoptic semantic map generation. OpenVSLAM~\cite{2019arXiv191001122S} was proposed by Sumikura~\textit{et al.}, which can adapt to panoramic cameras based on ORB-SLAM2~\cite{mur2017orb}. 
ROVINS~\cite{9144432} used four fisheye cameras and Inertial Measurement Units (IMUs) to achieve an omnidirectional multi-view VIO, which is robust in fast motion or sudden illumination changes.

Due to the low resolution of the fisheye lens at the edge of the FoV, many classic SLAM frameworks do not support the FoV from the negative half-plane beyond $180${\textdegree}, although they support fisheye camera models. 
However, PAL tends to maintain low radial distortion in the FoV from the negative half plane, which is comparable to the central FoV, as illustrated in Fig.~\ref{fig_wide_fov_cam}. 
Therefore, the mentioned classic SLAM frameworks above cannot adapt to the ultra-large FoV of PAL.
Chen~\textit{et al.}~\cite{chen2019palvo} proposed PALVO, introducing the Taylor camera model to VO methods and realizing high robustness in rapid motion and dynamic scenarios. 
PALVO is then updated to PA-SLAM~\cite{chen2021panoramic} with extra ORB features~\cite{2010BRIEF,2011ORB} extraction and the help of DBoW2~\cite{galvez2012bags} to incorporate a loop-closure module and become a SLAM system.
Hu~\textit{et al.}~\cite{hu2019indoor} proposed PAL-DSO, which extends the direct methods to PAL cameras. 
Wang~\textit{et al.}~\cite{wang2022pal} proposed PAL-SLAM to establish a complete visual SLAM framework based on PAL cameras and feature-based methods. 
Recently, Wang~\textit{et al.} updated their work to PAL-SLAM2~\cite{wang2024pal}, implementing support for IMU input. 

Wide-FoV input has been proven to be crucial for improving robustness and accuracy in positioning systems~\cite{wu2023_360vio,ahmadi2023hdpv,yang2024mcov}. 
Many extension works based on wide-FoV SLAM and visual odometry have appeared recently. 
Xie~\textit{et al.} implemented a lightweight panoramic depth completion network to acquire dense depth from a back-to-back dual fisheye camera setup and generate a dense 3D map~\cite{xie2024omnidirectional_dense}. 
Xiao~\textit{et al.} made use of panoramic semantic information to overcome the issue of inconsistent visual contents from different viewpoints in loop  closure~\cite{xiao2023semantic_loop_closure}. 

We have also made extensive explorations in the field of wide-FoV SLAM in the past. 
Building on the VINS series of works, we proposed LF-VIO~\cite{wang2022lf}, the first visual-inertial tightly coupled framework that can adapt to radial field-of-view inputs exceeding 180 degrees. 
Later, based on LF-VIO, we expanded it into a complete SLAM framework, LF-VISLAM~\cite{wang2023lfvislam}. 
Similarly, building on LF-VIO, we proposed LF-PGVIO~\cite{wang2023lfvislam,wang2024lfpgvio}, which introduced line features into the field of wide-field-of-view SLAM using geodesic constraints, achieving better positioning accuracy. 
In this paper, however, we focus on the impact of wide-field-of-view visual inputs on the construction of the SLAM problem itself, based on another equally mature SLAM framework, the monocular mode of ORB-SLAM3, and propose P2U-SLAM that can utilize point and pose uncertainties.

Although there have been plenty of VIO and SLAM frameworks that support wide-FoV cameras, almost all of them pay no attention to the uncertainty of variables observed multiple times by wide-FoV cameras and the corresponding variables' properties change during the location process. 
Some methods~\cite{qin2018vins,hu2019indoor,wang2022lf,matsuki2018omnidirectional} cannot update the uncertainty of variables continuously due to the limitation of direct methods or the optical flow tracking in the matching process. 
We point out that the methods based on descriptors have the potential to calculate and utilize the uncertainty of 3D points and 6D poses simultaneously. 
To make use of the potential of descriptors and overcome the limitations introduced by changes in variables' properties, we propose the first visual wide-FoV SLAM system utilizing point- and pose uncertainty based on ORB features.

\section{Camera Model}
\label{sec:camera_model}

The camera model describes the measurement process of the 3D coordinate of a map point and outputs a projection pixel on the image plane as its measurement result. 
Many SLAM systems rectify the images to adapt to the pinhole camera model, while the way of distortion rectification cannot be flexibly transferred to a wide-FoV camera, whose FoV can even surpass $180${\textdegree}.
Equidistance projection and other non-perspective projection camera models are more suitable for calibration of wide-FoV cameras~\cite{kannala2006generic} since the rectification of wide-FoV images is unnecessary.
Considering that the wide-FoV camera intrinsic parameters provided by different datasets may come from different camera models, we provide two popular camera models for wide-FoV cameras, which are the Taylor camera model~\cite{scaramuzza2006toolbox} and the Kannala-Brandt fisheye model~\cite{kannala2006generic}.

\subsection{Taylor Camera Model}
As described in~\cite{hu2019indoor,scaramuzza2006toolbox}, the projection function between a map point $\mathbf{p}$ with camera coordinate $(x,y,z)^T$ and its projection pixel $\mathbf{u}_\pi$ with coordinate $(u,v)^T$ can be written as
\begin{equation}
\label{Taylor_projection}
\mathbf{u}_\pi=\pi(\mathbf{p})=\rho(\theta) \cdot h(\mathbf{p})
+\begin{bmatrix} u_c \\ v_c \end{bmatrix},
\end{equation}
where
\begin{equation}
\label{Taylor_hp&rho&theta}
\begin{aligned}
    &h(\mathbf{p})=\begin{bmatrix} \frac{ x }{ \sqrt{x^2+y^2} }, & \frac{ y }{ \sqrt{x^2+y^2} } \end{bmatrix}^T,\\
    &\rho(\theta)=a_0+a_1\cdot\theta^1+a_2\cdot\theta^2+a_3\cdot\theta^3+...,\\
    &\theta=\arctan(\frac{ z }{ \sqrt{x^2+y^2} }).
\end{aligned}
\end{equation}
Meanwhile, the corresponding unprojection function is
\begin{equation}
\label{Taylor_unprojection}
\mathbf{\overline{p}}=\pi^{-1}(\mathbf{u}_\pi)=\frac{\mathbf{p}^{\prime}}{||\mathbf{p}^{\prime}||},
\end{equation}
where
\begin{equation}
\label{Taylor_unprojection_prime&zp&r}
\begin{aligned}
    &\mathbf{p}^{\prime}= \begin{bmatrix} u-u_c, & v-v_c, & z_p(r) \end{bmatrix}^T,\\
    &z_p(r)=b_0+b_1\cdot r^1+b_2\cdot\ r^2+b_3\cdot r^3+...,\\
    &r=\sqrt{(u-u_c)^2+(v-v_c)^2}.\\
\end{aligned}
\end{equation}
It can be observed that polynomials $\rho(\theta)$ and $z_p(r)$ 
and bias $\begin{bmatrix} u_c, v_c \end{bmatrix}^T$ are intrinsic parameters that need to calibrate.

It should be noted that the Jacobian matrix of the camera projection function can be calculated via chain rules as Eq.~\eqref{Taylor_Jacobian}, which matters in motion estimation and calculation of point uncertainty.
\begin{equation}
\label{Taylor_Jacobian}
    J_\pi=\frac{\partial\pi(\mathbf{p})}{\partial\mathbf{p}}
    =\frac{\partial\pi}{\partial\rho} \cdot \frac{\partial\rho}{\partial\theta} 
    \cdot \frac{\partial\theta}{\partial\mathbf{p}}
    +\frac{\partial\pi}{\partial h} \cdot \frac{\partial h}{\partial\mathbf{p}}.
\end{equation}

\subsection{Kannala-Brandt Fisheye Model}
The format of the Kannala-Brandt fisheye model (KB model)~\cite{kannala2006generic} is highly similar to the Taylor camera model, except that the KB model drops all the even terms in the polynomial of Eq.~\eqref{KB_projection} to get closer to the real physical model.
\begin{equation}
\label{KB_projection}
\rho(\theta)=k_1\cdot\theta^1+k_2\cdot\theta^3+k_3\cdot\theta^5+... .
\end{equation}
%

\section{Problem Statement}
\label{sec:problem_statement}

Many VO and visual SLAM methods use Bundle Adjustment (BA) to estimate the pose of sensor $\mathbf{T}_i$ at time $i$ and the position of the $j$th map point $\mathbf{p}_j$, where $ i=1,...,M $ and $ j=1,...,N $. 
These variables to estimate are represented as 
$x = \{\mathbf{T}_0,...,\mathbf{T}_M,\mathbf{p}_1,...,\mathbf{p}_N  \}$,
where one subset $\{\mathbf{T}_i,\mathbf{p}_j \} $ indicates a potential observation constraint between $i$th pose and $j$th map point. 
That constraint will be explicit when sensors receive measurement $\mathbf{u}_{ij}$ of $j$th map point at time $i$ and be expressed as 
\begin{equation}
\label{Observation}
    \mathbf{u}_{ij}=\pi(\mathbf{T}_i \cdot \mathbf{p}_j)+\mathbf{n}_{ij}, 
\end{equation}
where $\mathbf{n}_{ij}$ is random measurement noise and $\pi(\cdot)$ is measurement model of sensor such as Eq.~\eqref{Taylor_projection}.

The existing standard formulation of SLAM in practice derives from the essential paper of Lu and Milios~\cite{lu1997globally}, in which SLAM is summarized as a maximum likelihood estimation problem to obtain optimal variables to estimate, expressed as 
\begin{equation}
\label{MLE}
    \mathbf{x}^*=\mathop{\arg\max}\limits_{\mathbf{x}} {\rm P}(\mathbf{u}|\mathbf{x}),
\end{equation}
and when it is focused on one constraint between a single map point and the pose at a certain moment, the maximum likelihood estimation will be
\begin{equation}
\label{MLE_1C}
    \mathbf{T}_i^*,\mathbf{p}_j^*=\mathop{\arg\max}\limits_{\mathbf{T}_i,\mathbf{p}_j} {\rm P}(\mathbf{u}_{ij}|\mathbf{T}_i,\mathbf{p}_j),
\end{equation}
where the specific format of ${\rm P}(\mathbf{u}_{ij}|\mathbf{T}_i,\mathbf{p}_j)$ is determined by the probability distribution of the sensor's measurement noise $\mathbf{n}_{ij}$ in Eq.~\eqref{Observation}, which is usually assumed as a zero-mean Gaussian distribution with covariance matrix ${\Sigma}_{ij}$. Then, Eq.~\eqref{MLE_1C} can be rewritten as
\begin{equation}
\begin{aligned}
\label{MLE_1C_GUASS}
    \mathbf{T}_i^*,\mathbf{p}_j^*
    &=\mathop{\arg\min}\limits_{\mathbf{T}_i,\mathbf{p}_j} 
    -{\rm log}({\rm P}(\mathbf{u}_{ij}|\mathbf{T}_i,\mathbf{p}_j))  \\
    &=\mathop{\arg\min}\limits_{\mathbf{T}_i,\mathbf{p}_j} 
    ||\pi(\mathbf{T}_i \cdot \mathbf{p}_j) - \mathbf{u}_{ij}||_{{\Sigma}_{ij}^{-1}}^2,
\end{aligned}
\end{equation}
where the notion $||e||_{{\Sigma}^{-1}}^2=e^T{\Sigma}^{-1}e$. 

It should be noted that in Eq.\eqref{MLE_1C_GUASS}, both $\mathbf{T}_i$ and $\mathbf{p}_j$ are the targets we aim to solve for, and only $\mathbf{u}_{ij}$ represents the observed results. 
The original intention of Eq.\eqref{MLE_1C_GUASS} is to find a set of $\mathbf{T}_i$ and $\mathbf{p}_j$ such that the likelihood probability is maximized.
Therefore, the covariance ${\Sigma}_{ij}$ of the likelihood distribution is only related to the observed result $\mathbf{u}_{ij}$, and is independent of $\mathbf{T}_i$ and $\mathbf{p}_j$.
Meanwhile, the result of Eq.\eqref{MLE_1C_GUASS} is exactly the optimization objective function for global Bundle Adjustment (BA) in the visual SLAM problem.

\subsection{Point Uncertainty}
This subsection introduces the role that 3D point uncertainty plays in motion estimation and explains how it influences estimation.

Eq.~\eqref{MLE_1C_GUASS} works well in a situation where all the variables to estimate are involved in the BA process while existing SLAM and VO methods utilize the technique of local map optimization considering the cumulative quantity of variables along with time and restrictive computing resources. 
The application of a local map limits the number of variables involved in BA to ensure the efficiency of optimization. 
In the meantime, it introduces the transformation of partial variables' properties, especially when the sensor is a fisheye or a PAL camera owning an ultra-wide overlap area, which past wide-FoV SLAM frameworks ignore.
The much wider overlap area between frames increases the probability that the map points generated in the past are observed by the newest frame. 
Thus, the wide FoV results in a greater proportion of well-estimated past map points in the local map. 
In order to avoid computational costs and help estimate the pose of the latest frame, these past map points are treated as measurement results rather than estimation variables.
This shift in data properties has become more common with the introduction of wide-FoV sensors.
Therefore, the uncertainty of these past map points should be introduced and the conditional probability part in Eq.~\eqref{MLE_1C} also needs to be modified. 
Considering a observation constraint between a pose to estimate $\mathbf{T}_i$ and a past map point $\mathbf{p}_p$, we have
\begin{equation}
\label{MLE_1C_ONLYPOSE}
    \mathbf{T}_i^*=\mathop{\arg\max}\limits_{\mathbf{T}_i} 
{\rm P}(\mathbf{u}_{ip},\mathbf{p}_p|\mathbf{T}_i),
\end{equation}
where the data property of $\mathbf{p}_p$ is converted from the variable to be estimated in Eq.\eqref{MLE_1C_GUASS} into observable data. There are two measurements. One is $\mathbf{u}_{ip}$ with random noise $\mathbf{n}_{ip} \sim N(0,{\Sigma}_{ip})$ and the other is $\mathbf{p}_p\sim N(\hat{\mathbf{p}_p}, {\Sigma}_{3\times3,p})$. ${\Sigma}_{3\times3,p}$ denotes the the uncertainty of past map point $\mathbf{p}_p$. 
Then, the covariance ${\Sigma}_{ip}^{\prime}$ of ${\rm P}(\mathbf{u}_{ip},\mathbf{p}_p|\mathbf{T}_i)$ can be calculated according to Eq.~\eqref{Taylor_Jacobian} and Eq.~\eqref{Observation}:
\begin{equation}
\label{Covariance_Point}
    {\Sigma}_{ip}^{\prime}= 
    \nabla_{\hat{\mathbf{p}_p}}\pi \cdot {\Sigma}_{3\times3,p} \cdot 
    (\nabla_{\hat{\mathbf{p}_p}}\pi)^T +{\Sigma}_{ip}, \\
\end{equation}
where
\begin{equation}
\begin{aligned}
\label{Jacobian_Point_simplification}
    \nabla_{\hat{\mathbf{p}_p}}\pi
    &=\frac{\partial\pi(\mathbf{T}_i\cdot\hat{\mathbf{p}_p})}{\partial\hat{\mathbf{p}_p}}\\
    &=\frac{\partial\pi(\mathbf{T}_i\cdot\hat{\mathbf{p}_p})}{\partial(\mathbf{T}_i\cdot\hat{\mathbf{p}_p})} 
    \cdot \frac{\partial(\mathbf{T}_i\cdot\hat{\mathbf{p}_p})}{\partial\hat{\mathbf{p}_p}}\\
    &= J_{\pi(\mathbf{T}_i\cdot\hat{\mathbf{p}_p})}\cdot \mathbf{R}_{i}.
\end{aligned}
\end{equation}
$\mathbf{R}_{i}$ denotes the rotation matrix at time $i$.
Therefore, Eq.~\eqref{MLE_1C_GUASS} finally becomes
\begin{equation}
\label{MLE_1C_ONLYPOSE_GUASS}
\begin{aligned}
    \mathbf{T}_i^*
    &=\mathop{\arg\min}\limits_{\mathbf{T}_i} 
    -{\rm log}({\rm P}(\mathbf{u}_{ip},\mathbf{p}_p|\mathbf{T}_i))  \\
    &=\mathop{\arg\min}\limits_{\mathbf{T}_i} 
    ||\pi(\mathbf{T}_i \cdot \hat{\mathbf{p}_p}) - \mathbf{u}_{ip}||_{{\Sigma}_{ip}^{\prime-1}}^2.
\end{aligned}
\end{equation}
The point uncertainty ${\Sigma}_{3\times3,p}$ is determined by historical observation of map point $\mathbf{p}_p$. 
For one historical observation $\mathbf{u}_{kp}$ at time $k$, the 3D observation error $\mathbf{r}_{kp}$ of $\mathbf{p}_p$ is defined as
\begin{equation}
\label{Observation_error}
    \mathbf{r}_{kp}=\mathbf{R}_k^T \cdot \changedROne{(||\mathbf{T}_k \cdot \mathbf{p}_p|| \cdot {\pi}^{-1}(\mathbf{u}_{kp})
    - \mathbf{T}_k \cdot \mathbf{p}_p)}, 
\end{equation}
and the point uncertainty ${\Sigma}_{3\times3,p}$ is
\begin{equation}
\label{point_uncertainty}
    {\Sigma}_{3\times3,p}=\frac{ 1 }{ N_K-1 }\sum_{k\in K}\mathbf{r}_{kp} \cdot {\mathbf{r}_{kp}}^{T}, 
\end{equation}
where $K$ denotes a subset of all the timestamps of the map point $\mathbf{p}_p$ historical observation, and $N_K$ denotes the count of historical observations.

\subsection{Pose Uncertainty}
Similar to point uncertainty, pose uncertainty is introduced when an observation constraint is established between a fully optimized pose and a newly added map point. That past pose becomes the measurement result, while the newly added map point is the target to estimate. 
Considering a observation constraint between a past pose $\mathbf{T}_p$ and a map point to estimate $\mathbf{p}_j$, we have
\begin{equation}
\label{MLE_1C_ONLYPOINT}
\begin{aligned}
    \mathbf{p}_j^*
    &=\mathop{\arg\max}\limits_{\mathbf{p}_j} 
    {\rm P}(\mathbf{u}_{pj},\mathbf{T}_p|\mathbf{p}_j), \\
    &=\mathop{\arg\min}\limits_{\mathbf{p}_j} 
    -{\rm log}({\rm P}(\mathbf{u}_{pj},\mathbf{T}_p|\mathbf{p}_j)  \\
    &=\mathop{\arg\min}\limits_{\mathbf{p}_j} 
    ||\pi(\mathbf{T}_p \cdot \hat{\mathbf{p}_j}) - \mathbf{u}_{pj}||_{{\Sigma}_{pj}^{\prime\prime-1}}^2.
\end{aligned}
\end{equation}
\addROne{where the data property of $\mathbf{T}_p$ is converted from the variable to be estimated in Eq.\eqref{MLE_1C_GUASS} into observable data.
The distribution of the pose measurement values is represented in the form of Lie algebra, which is $\xi_p\sim N(\hat{\xi_p}, {\Sigma}_{6\times6,p})$.
${\Sigma}_{6\times6,p}$ denotes the the uncertainty of past pose's Lie algebra $\xi_p$
.}
And the corresponding covariance ${\Sigma}_{pj}^{\prime\prime}$ is
\begin{equation}
\label{Covariance_POSE}
    {\Sigma}_{pj}^{\prime\prime}=
    \nabla_{\hat{\xi_p}}\pi \cdot {\Sigma}_{6\times6,p} \cdot 
    (\nabla_{\hat{\xi_p}}\pi)^T +{\Sigma}_{pj}, \\
\end{equation}
where
\begin{equation}
\begin{aligned}
\label{Jacobian_Pose_simplification}
    \nabla_{\hat{\xi_p}}\pi
    &=\frac{\partial\pi(\mathbf{T}_p\cdot\mathbf{p}_j)}{\partial\delta\xi}\\
    &=\frac{\partial\pi(\mathbf{T}_p\cdot\mathbf{p}_j)}{\partial(\mathbf{T}_p\cdot\mathbf{p}_j)} 
    \cdot \frac{\partial(\mathbf{T}_p\cdot\mathbf{p}_j)}{\partial\delta\xi}\\
    &= J_{\pi(\mathbf{T}_p\cdot\mathbf{p}_j)}
    \cdot \begin{bmatrix} -(\mathbf{T}_p\cdot\mathbf{p}_j)^\wedge & I_{3\times3}\end{bmatrix},
\end{aligned}
\end{equation}
which applies a perturbation model~\cite{1972Introduction} and involves the exponential mapping of the Lie group, whose specific form is as follows,
\begin{equation} 
\begin{aligned}
\label{Li_exponential_mapping}
    \mathbf{T}_p = {\rm exp}(\hat{\xi_p}^\wedge).
\end{aligned}
\end{equation}
Similarly to Eq.~\eqref{Observation_error}, the pose uncertainty ${\Sigma}_{6\times6,p}$ is determined by the mapping of the two-dimensional historical observations in the Lie algebra space of the pose.
\addROne{Unlike the calculation of 3D map point observation error, where we have a camera back-projection function that can directly obtain 3D observation results from 2D observations to simplify the calculation,
in the acquisition of pose observation error, there is no such convenient function for converting between 2D point observations and 6D pose observations.
Therefore, here we directly use the pseudo-inverse of the gradient matrix of the 2D projection with respect to the pose Lie algebra and the 2D observation error to calculate the pose observation error.}
For one historical observation $\mathbf{u}_{ph}$ at time $p$, the 6D observation error $\mathbf{r}_{ph}$ of $\mathbf{\xi}_p$ is defined as
\begin{equation} 
\begin{aligned}
\label{Observation_error_pose}
    \mathbf{r}_{ph}=(\nabla_{\hat{\xi_p}}\pi(\mathbf{T}_p\cdot\mathbf{p}_h))^{-1} \cdot 
    (\mathbf{u}_{ph} - {\pi}(\mathbf{T}_p \cdot \mathbf{p}_h)), 
\end{aligned} 
\end{equation}
where $(\nabla_{\hat{\xi_p}}\pi)^{-1}$ denotes the pseudo-inverse of $\nabla_{\hat{\xi_p}}\pi$
\begin{equation} 
\begin{aligned}
\label{Pseudo_inverse}
    (\nabla_{\hat{\xi_p}}\pi)^{-1}=
    (\nabla_{\hat{\xi_p}}\pi)^T \cdot 
    (\nabla_{\hat{\xi_p}}\pi \cdot (\nabla_{\hat{\xi_p}}\pi)^T)^{-1}, 
\end{aligned} 
\end{equation}
and the pose uncertainty ${\Sigma}_{6\times6,p}$ is
\begin{equation}
\label{pose_uncertainty}
    {\Sigma}_{6\times6,p}=\frac{ 1 }{ N_H-1 }\sum_{h\in H}\mathbf{r}_{ph} \cdot {\mathbf{r}_{ph}}^{T}, 
\end{equation}
where $H$ denotes a subset of all the historical observations $\{\mathbf{u}_{ph}|h \in H\}$ in keyframe at time $p$ whose pose is $\mathbf{T}_p$, 
and $N_H$ denotes the count of historical observations.

Overall, point uncertainty and pose uncertainty will be used to reduce errors introduced by treating past points and poses as measurement results in the estimation process. Their specific application within the SLAM framework will be detailed in the next section.

\newpage
\section{System Overview}
\label{sec:system_overview}

\begin{figure*}[t]
    \centering
    \includegraphics[width=5.0in]{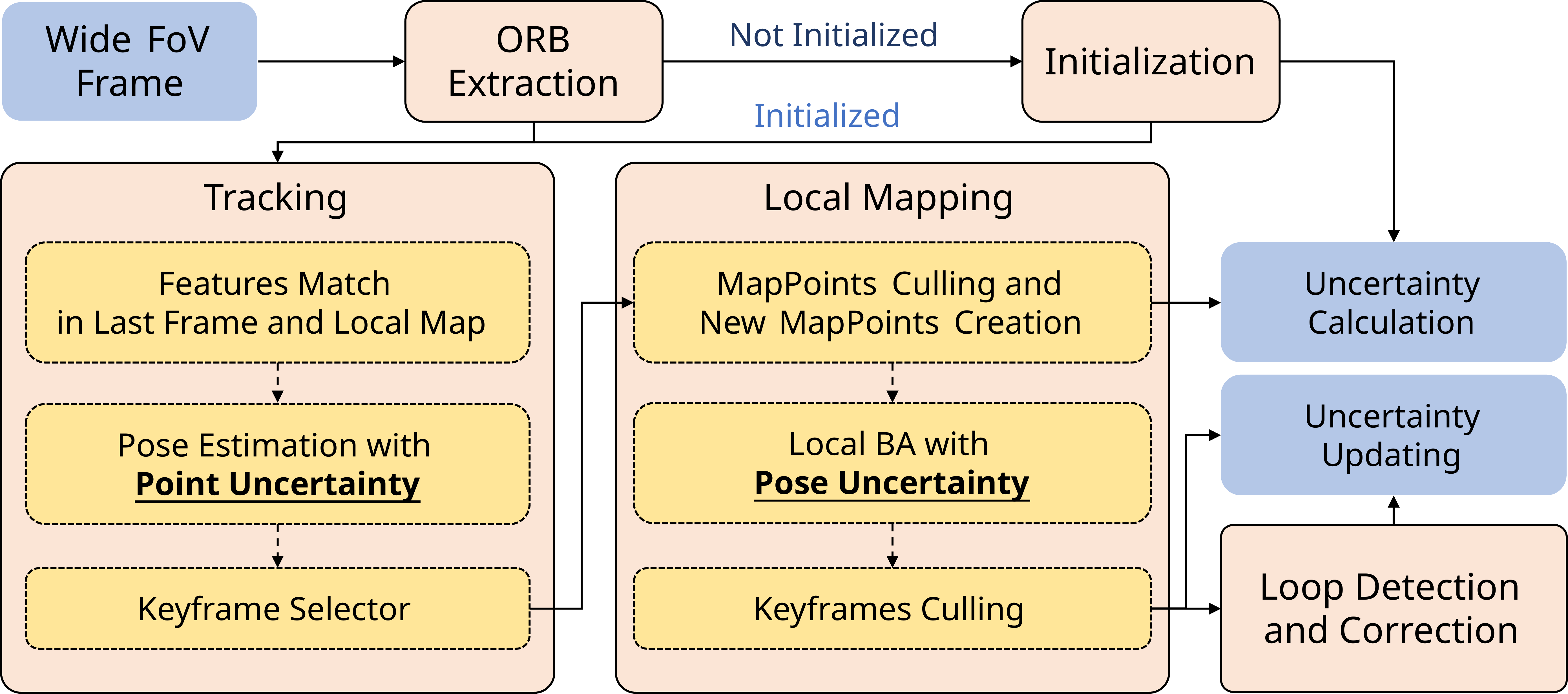}
    \caption{The pipeline of P2U-SLAM. 
    P2U-SLAM mainly consists of initialization, tracking, local BA, and loop closing. Point uncertainty functions in the tracking module to suppress the noise from treating past map points as measurement results on the current frame's pose estimation. 
    Pose uncertainty acts on the local BA module to suppress the noise from treating fixed keyframe poses as measurement results on other variables to estimate.
    }
    \label{fig_pipline}
    \end{figure*}

A pipeline of our system is schematically given as Fig.~\ref{fig_pipline}. 
P2U-SLAM is established on the monocular mode of ORB-SLAM3~\cite{campos2021orb} and mainly includes initialization, tracking, local bundle adjustment, and loop closing. 
The construction, application, and maintenance of point and pose uncertainties have been integrated into the main framework of P2U-SLAM to provide robustness and positioning accuracy.
And ORB features are selected to apply in the feature matching of the whole above process.

\subsection{Initialization}
As stated in PALVO~\cite{chen2019palvo}, the map points observed by the wide-FoV camera are barely possible to be coplanar. 
Thus essential matrix is utilized to resolve the process of structure from motion (SFM). The essential matrix satisfies the epipolar constraints.
\begin{equation}
\label{SFM_E}
    (\mathbf{\overline{p}}^{i2})^T(\mathbf{t}^\wedge{}\mathbf{R})\mathbf{\overline{p}}^{i1}=0, 
\end{equation}
where $\mathbf{t}^\wedge{}\mathbf{R}$ denotes the essential matrix and $\mathbf{\overline{p}}^i$ denotes the unit vector of a map point in the camera coordinate at time $i$, and can be calculated by an unprojection function mentioned in Eq.~\eqref{Taylor_unprojection}. 

The eight-point algorithm with a RANSAC scheme~\cite{longuet1981computer} is applied to exclude outliers of keypoint matches. 
However, the epipolar in the non-perspective projection camera model as mentioned in Sec.~\ref{sec:camera_model} is not a straight line but a ``curve'', which makes it difficult to calculate reprojection errors in polar geometry. 
Therefore, the reprojection error is redesigned as the reprojection of vertical distance between $\mathbf{\overline{p}}^{i2}$ and the polar plane, which is used to be the inlier-checking metric of RANSAC.
After successful initialization, the system will begin to estimate the pose using the constraint relationship between map points and the current frame.

\begin{figure}[h]
    \centering    
    \subfigure[]{
        \label{fig:tracing_uncertainty}
        \begin{minipage}[t]{0.9\linewidth}
            \centering
            \includegraphics[width=1\linewidth]{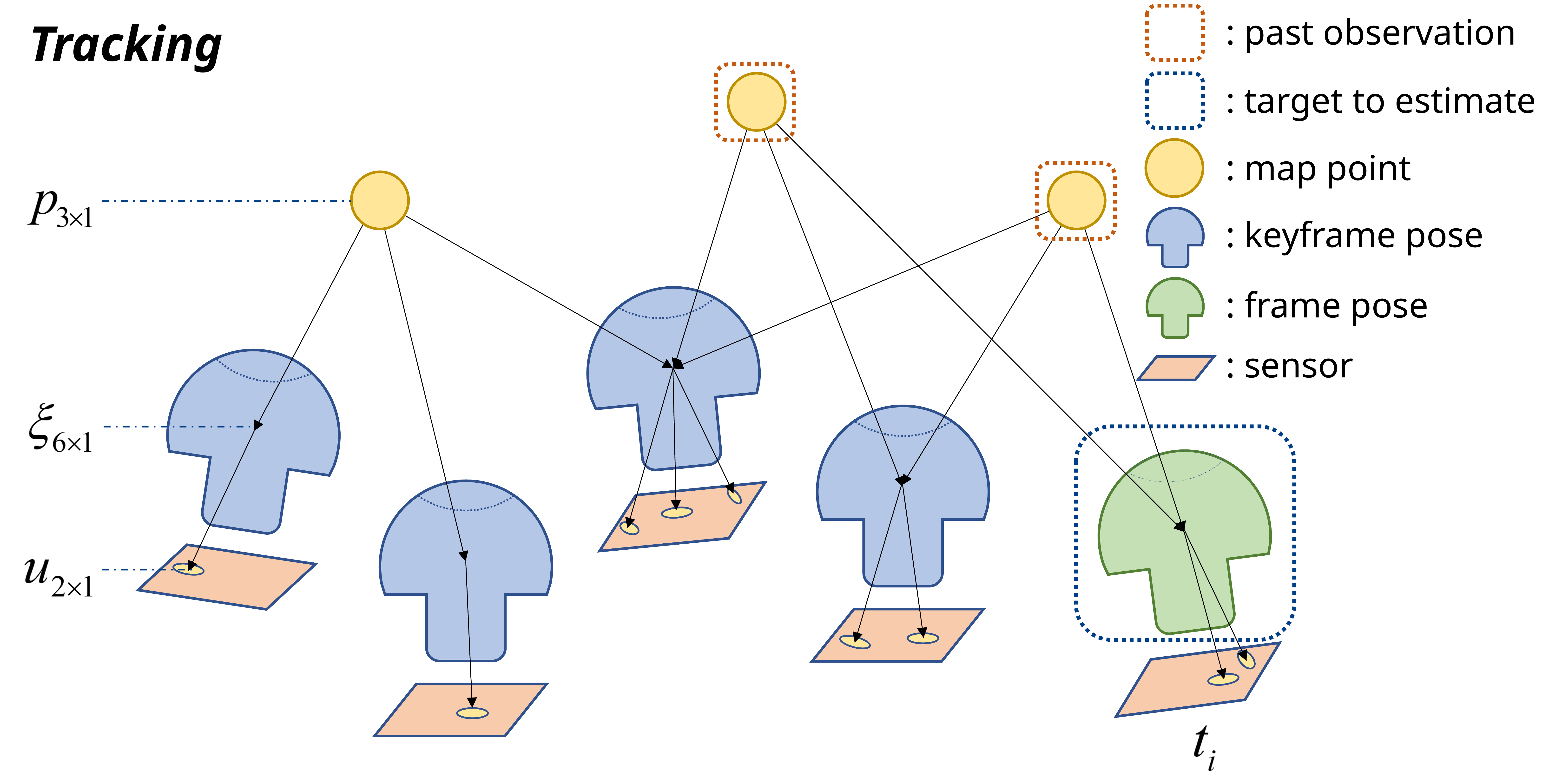}
        \end{minipage}
    }
    

    \subfigure[]{
        \label{fig:lba_uncertainty}
        \begin{minipage}[t]{0.9\linewidth}
            \centering
            \includegraphics[width=1\linewidth]{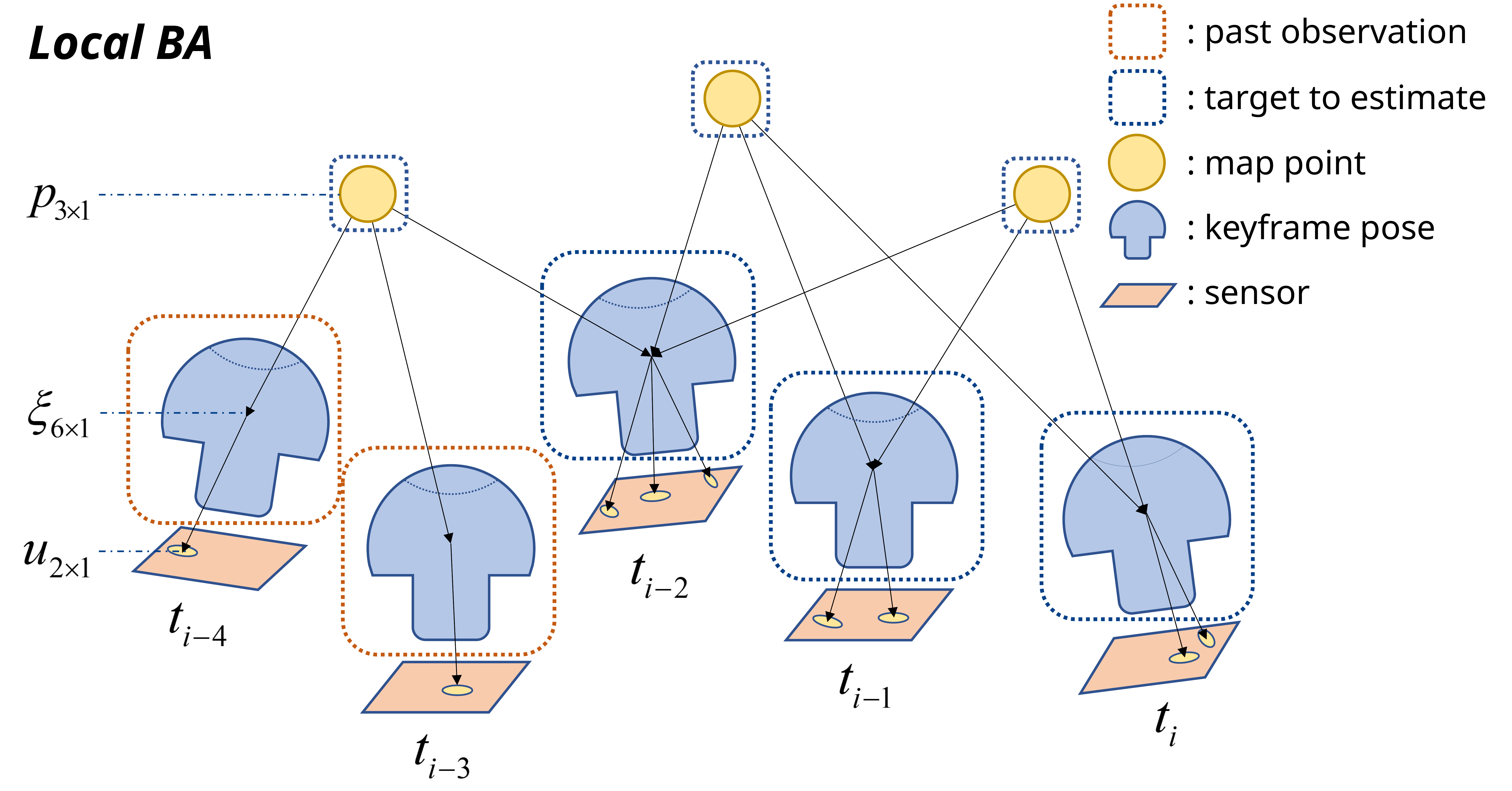}
        \end{minipage}
    }
    
    \caption{Illustrations of situations in which point and pose uncertainty are applied.
        (a) The map points are viewed as fixed when estimating the pose of a new frame of tracking, so the point uncertainty should be applied according to Eq.~\eqref{Covariance_Point} and Eq.~\eqref{MLE_1C_ONLYPOSE_GUASS}. 
        (b) In the process of local BA, there are some fixed keyframes viewed as observation results that are not targets to estimate while still having a great influence on optimization. Similarly, the pose uncertainty of these fixed keyframes (with orange dotted boxes) should be processed as Eq.~\eqref{MLE_1C_ONLYPOINT} and Eq.~\eqref{Covariance_POSE}.}
    \label{fig:uncertainty}
\end{figure}

\subsection{Tracking with Point Uncertainty}
The tracking module aims at real-time updating of the agent's pose. 
In this module, as shown in Fig.~\ref{fig:tracing_uncertainty}, the only variable to estimate is the pose $\mathbf{T}_i$ of the new frame and all of the map points $\mathbf{p}_j$ with matched 2D-features $\mathbf{u}_{ij}$ are measurement results, so according to Eq.~\eqref{MLE_1C_ONLYPOSE_GUASS}, we have 
\begin{equation}
\label{MLE_NC_ONLYPOSE_GUASS}
    \mathbf{T}_i^*
    =\mathop{\arg\min}\limits_{\mathbf{T}_i} \sum_{j\in E}
    ||\pi(\mathbf{T}_i \cdot \hat{\mathbf{p}_j}) - \mathbf{u}_{ij}||_{{\Sigma}_{ij}^{\prime-1}}^2,
\end{equation}
where $E$ is the set of map points observed in the new frame, and point uncertainty influences the direction of optimization by acting as the information matrix ${\Sigma}_{ij}^{\prime-1}$ in Eq.~\eqref{Covariance_Point}.
The tracking module is where real-time performance truly shines, and with point uncertainty, P2U-SLAM enhances the accuracy of pose estimation. This approach allows for better handling of past map points as measurements, leading to more precise tracking.

\begin{figure*}[b]
    \centering
    
    \subfigure{
        \rotatebox{90}{\scriptsize{~~~~~~~~~~~~~Seq. ID02}}
        \begin{minipage}[t]{0.47\linewidth}
            \centering
            \includegraphics[width=1\linewidth]{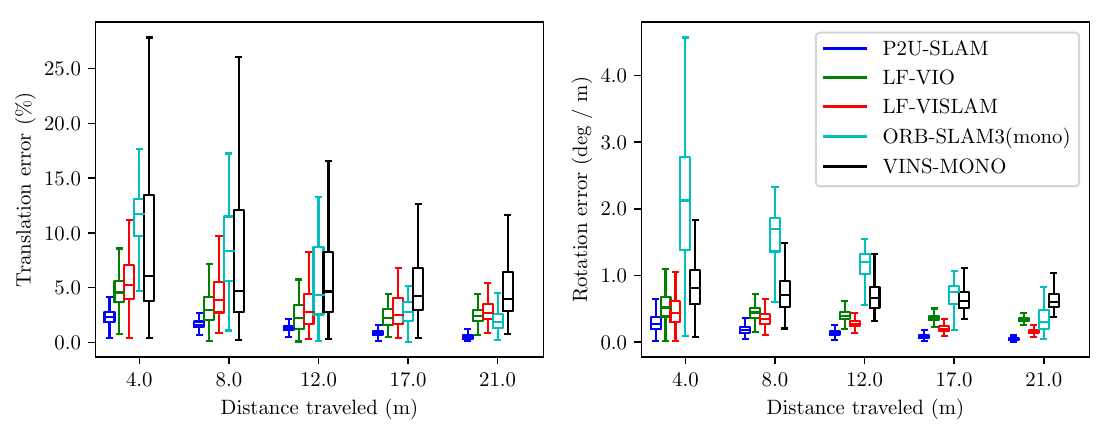}
        \end{minipage}
    }
    \subfigure{
        \begin{minipage}[t]{0.47\linewidth}
            \centering
            \includegraphics[width=1\linewidth]{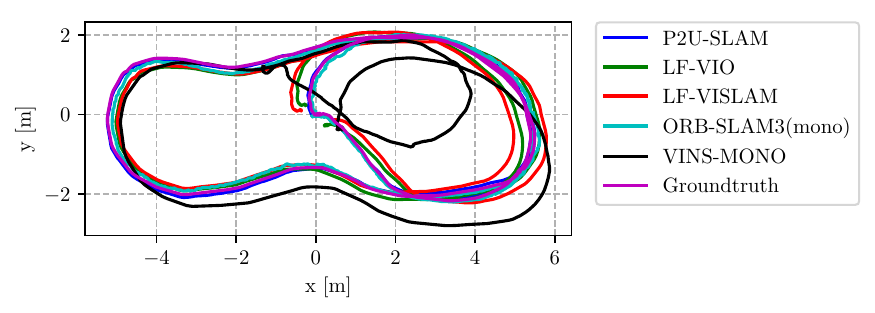}
        \end{minipage}
    }
    \vspace{-3mm}
    \setcounter{subfigure}{0}
    
    \subfigure{
        \rotatebox{90}{\scriptsize{~~~~~~~~~~~~~Seq. ID04}}
        \begin{minipage}[t]{0.47\linewidth}
            \centering
            \includegraphics[width=1\linewidth]{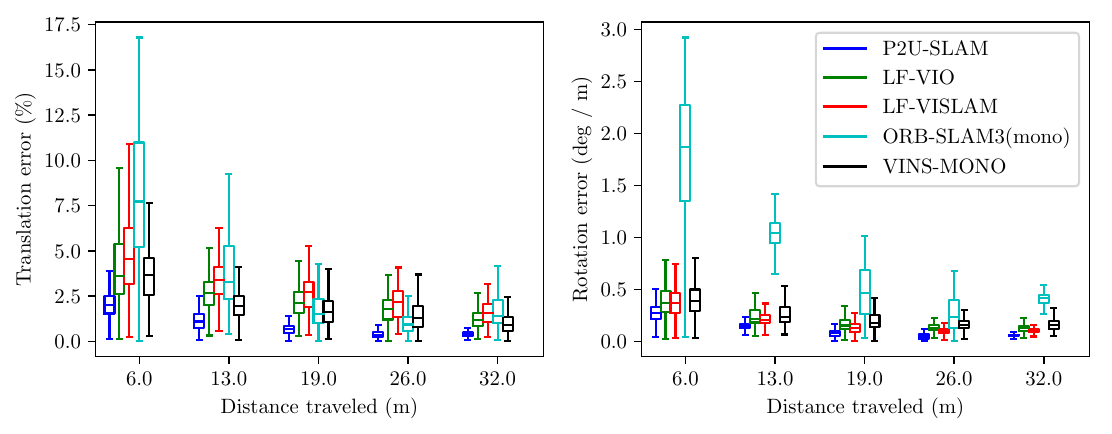}
        \end{minipage}
    }
    \subfigure{
        \begin{minipage}[t]{0.47\linewidth}
            \centering
            \includegraphics[width=1\linewidth]{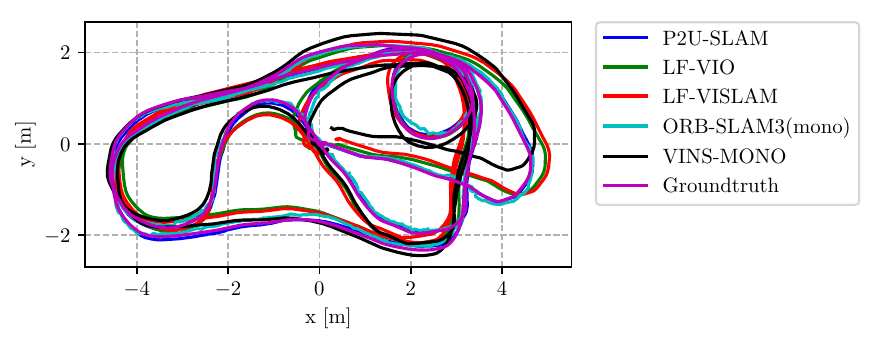}
        \end{minipage}
    }
    \vspace{-3mm}
    \setcounter{subfigure}{0}
    
    \subfigure[Translation and  Rotation Error]{
        \rotatebox{90}{\scriptsize{~~~~~~~~~~~~~Seq. ID08}}
        \begin{minipage}[t]{0.47\linewidth}
            \centering
            \includegraphics[width=1\linewidth]{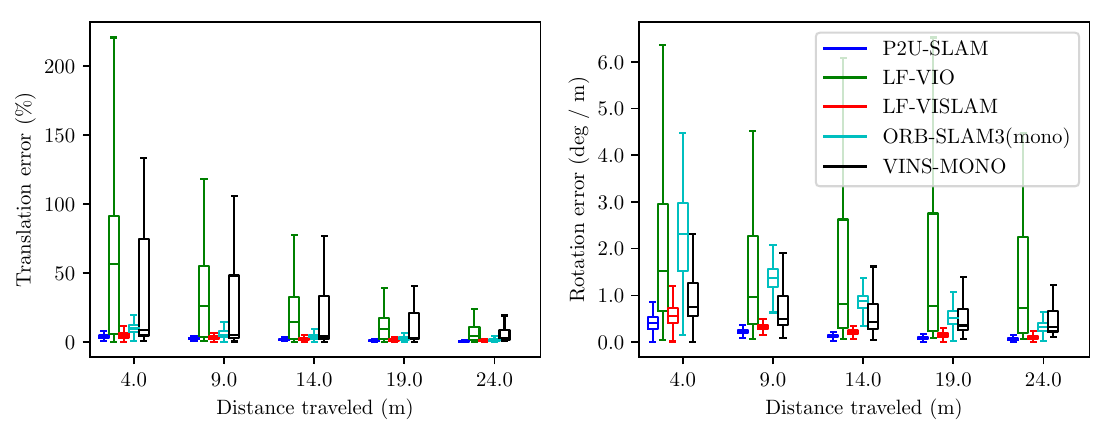}
        \end{minipage}
    }
    \subfigure[Top Trajectory]{
        \begin{minipage}[t]{0.47\linewidth}
            \centering
            \includegraphics[width=1\linewidth]{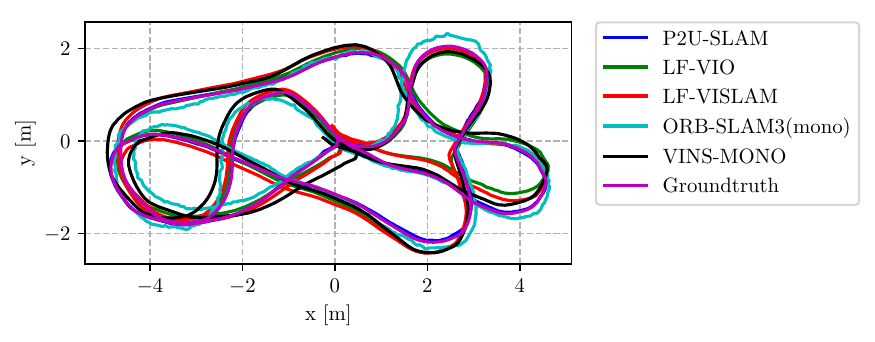}
        \end{minipage}
    }
    \caption{Examples of error analyses and top trajectories of different SLAM and VIO systems on the PALVIO dataset~\cite{wang2022lf}. 
    The details of the a) Translation and Rotation Error subfigures illustrate that P2U-SLAM not only has a low average error but also exhibits a more concentrated and stable error distribution. 
    In the b) Top Trajectory subfigures, the clear superiority of P2U-SLAM in overlap with the ground truth trajectory over VINS-Mono and LF-VIO demonstrates its advantage of low absolute trajectory error, while its higher trajectory smoothness compared to ORB-SLAM3 indicates the stability of its real-time tracking. 
    This is attributed to the innovative integration of point- and pose uncertainty within the P2U-SLAM framework, which effectively mitigates the impact of weak feature correspondences and enhances the overall reliability of the system.}
    \label{fig:palvio_dataset_trans_rot_error}
\end{figure*}

\subsection{Local BA with Pose Uncertainty}
The local BA module further optimizes the pose of local keyframes and all visible map points to provide reliable visual constraints for subsequent tracking.
This means that the pose of local keyframes and all visible map points are our optimization targets, and their uncertainties are exactly what we aim to reduce. 
Therefore, those uncertainties cannot be used as the covariance of observed data to participate in the optimization.
That is to say, only the uncertainty of the pose of the fixed keyframe can be used for Local BA.
In the process of local BA, the whole map points are set variables to estimate, while keyframes are divided into fixed keyframes (observation results) and local keyframes (variables to estimate) as shown in Fig.~\ref{fig:lba_uncertainty}. 
Therefore, the optimization can be expressed as 
\begin{equation}
\label{MLE_NC_PARTPOSE_GUASS}
\begin{aligned}
    \mathbf{T}_{I_1}^*, \mathbf{p}^*
    =\mathop{\arg\min}\limits_{\mathbf{T}_{I_1},\mathbf{p}} \sum_{j}
    (&\sum_{i_1\in I_1}
    ||\pi(\mathbf{T}_{i_1} \cdot \mathbf{p}_j) - \mathbf{u}_{i_1j}||_{{\Sigma}_{i_1j}^{-1}}^2 \\
    +&\sum_{i_2\in I_2}
    ||\pi(\mathbf{T}_{i_2} \cdot \mathbf{p}_j) - \mathbf{u}_{i_2j}||_{{\Sigma}_{i_2j}^{\prime\prime-1}}^2) ,
\end{aligned}
\end{equation}
where $I_1$ denotes the set of local keyframes' timestamps and $I_2$ denotes the set of fixed keyframes' timestamps. 
Pose uncertainty is expressed as information matrix ${\Sigma}_{i_2j}^{\prime\prime-1}$ in Eq.~\eqref{Covariance_POSE} to adjust the weights for optimization.
Local BA plays a significant role in refining the estimates, and by incorporating pose uncertainty, P2U-SLAM achieves even greater optimization accuracy. This method ensures that fixed keyframes, treated as observation results, have their uncertainties properly accounted for in the optimization process.

\subsection{Loop Closing}
In the thread of loop closing, all ORB features obtained from keyframes have been transformed to BoW vectors with DBoW2~\cite{galvez2012bags}. 
The candidate loop keyframes are filtered by comparing BoW similarity with the current keyframe. 
False positives in loop detection tend to appear more frequently with a wide-FoV camera because of its much bigger overlapping area. 
Therefore, the epipolar constraints mentioned in Eq.~\eqref{SFM_E} will also be utilized to check between the candidate loop keyframe and the current keyframe. 
Once a true positive loop closure is detected, the relative pose of the current keyframe from the candidate loop keyframe will be calculated and all uncertainty of poses and points will be updated. 
Finally, a global BA is processed to optimize keyframe poses and map points.

%

\newpage
\section{Experiments}
\label{sec:experiments}
In this section, the proposed P2U-SLAM is evaluated on two public datasets, including the PALVIO dataset~\cite{wang2022lf} and the TUM-VI benchmark~\cite{schubert2018tum}. 
The PALVIO dataset, recorded with two Panoramic Annular Lens cameras and an IMU sensor, is collected on an aerial vehicle as shown in Fig.~\ref{fig:exibition_for_mobile_platform}, providing a comprehensive dataset for assessing SLAM in indoor environments.
The TUM-VI benchmark, on the other hand, features fisheye inputs that mimic the viewpoint of a vehicle or drone in motion, offering a set of challenging sequences to test the robustness of SLAM systems in various dynamic conditions.
Notably, P2U-SLAM is a monocular visual SLAM system, and all the compared methods are set to the monocular mode for a fair comparison except that IMU data input is imperative. 
The estimated trajectories of SLAM and VIO systems are aligned with the ground truth using a Sim(3) transformation. 
All experiments have been run on an AMD Core r5-4650g CPU, with 16GB memory, using only the CPU for computing.

\begin{table*}[b] 
	\caption{Performance comparison on the public PALVIO dataset~\cite{wang2022lf} (ATE in $m$, RPEt in $\%$, RPEr in $degree/m$)$^1$.}        
	\label{EVAL_on_PALVIO_DATASET}           
    \begin{threeparttable}
	\begin{center}                 
            \renewcommand\arraystretch{1.2}{\setlength{\tabcolsep}{2.8mm}{\begin{tabular}{c c c c c c c c c c c c c}
			\toprule               
                \multicolumn{2}{c}{\multirow{2}{*}{SLAM/VIO Method}} & \multicolumn{10}{c}{Sequences}                                                                                                                                          \\
                \multicolumn{2}{c}{} & ID01           & ID02           & ID03           & ID04           & ID05          & ID06           
                                     & ID07           & ID08           & ID09           & ID10           &Avg$^2$  \\

                \midrule 
                \multirow{3}{*}{SVO2.0~\cite{forster2016svo}$^3$}      
                    & RPEt           &6.531           &6.995           &2.710           &1.928           &2.354          &3.409           
                                     &3.718           &2.811           &2.012           &14.147          &4.662 \\
                    & RPEr           &0.401           &\textbf{0.378}  &0.235           &0.165           &0.296          &0.320           
                                     &0.187           &\textbf{0.291}  &\textbf{0.230}  &0.608           &0.311 \\
                    & ATE            &0.761           &0.380           &0.366           &0.174           &0.148          &0.124           
                                     &0.428           &0.236           &0.292           &1.122           &0.403 \\
                \midrule 
                \multirow{3}{*}{VINS-Mono~\cite{qin2018vins}}      
                    & RPEt           &4.118           &5.936           &2.068           &\underline{1.098}&/             &2.231           
                                     &/               &3.567           &9.747           &\underline{2.372}&3.114* \\
                    & RPEr           &0.423           &0.919           &0.225           &0.165           &/              &0.218           
                                     &/               &0.478           &2.058           &\underline{0.284}           &0.477* \\
                    & ATE            &0.449           &0.783           &0.243           &0.287           &/              &0.121           
                                     &/               &0.516           &1.463           &\underline{0.216}&0.408* \\
			\midrule
                \multirow{3}{*}{LF-VIO~\cite{wang2022lf}}      
                    & RPEt           &\underline{3.595}&2.601          &2.136           &1.315           &6.362          &\underline{1.504}           
                                     &9.699           &8.225           &12.581          &2.566           &5.058 \\
                    & RPEr           &\textbf{0.367}  &0.809           &0.202           &0.132           &0.668          &0.190           
                                     &1.603           &2.278           &1.420           &0.347           &0.802 \\
                    & ATE            &0.400           &0.286           &0.270           &0.284           &2.199          &0.111           
                                     &2.070           &2.090           &2.360           &0.225           &1.030 \\
                \midrule 
                \multirow{3}{*}{\add{LF-VISLAM}~\cite{wang2023lfvislam}}      
                    & \add{RPEt}     &\add{6.869}     &\add{3.026}     &\add{2.621}     &\add{1.707}     &\add{\underline{1.448}} &\add{1.687}
                                     &\add{\underline{1.479}}&\add{\underline{1.835}}&\add{\underline{1.516}}&\add{3.493}&\add{2.568}\\
                    & \add{RPEr}     &\add{0.398}     &\add{0.740}     &\add{\underline{0.199}}&\add{\underline{0.111}}&\add{\underline{0.139}}  &\add{\underline{0.171}}    
                                     &\add{\underline{0.145}}     &\add{0.348}     &\add{0.362}     &\add{0.300}     &\add{0.291}\\
                    & \add{ATE}      &\add{\underline{0.360}}&\add{0.398}&\add{0.460}   &\add{0.276}     &\add{0.181}    &\add{\underline{0.102}}      
                                     &\add{\underline{0.263}}&\add{\underline{0.190}}&\add{\underline{0.171}}&\add{0.339}&\add{0.274}\\
                \midrule
                \multirow{3}{*}{ORB-SLAM3 (mono)~\cite{campos2021orb}}      
                    & RPEt           &/               &\underline{2.132}&\underline{1.604}&1.935         &1.861          &2.009           
                                     &2.010           &2.239           &2.662           &/               &1.645* \\
                    & RPEr           &/               &0.795           &0.356           &0.408           &0.390          &0.532           
                                     &0.357           &0.451           &0.607           &/               &0.390* \\
                    & ATE            &/               &\underline{0.127}&\underline{0.098}&\underline{0.095}&\underline{0.105}&0.111           
                                     &0.264           &0.271           &0.520           &/               &0.159* \\
			\midrule
                \multirow{3}{*}{P2U-SLAM (ours)}      
                    & RPEt           &\textbf{3.388}  &\textbf{0.567}  &\textbf{0.96}   &\textbf{0.452}  &\textbf{0.546} &\textbf{0.898}  
                                     &\textbf{0.505}  &\textbf{1.301}  &\textbf{0.866}  &\textbf{1.123}  
                                     &\textbf{1.061} \\
                    & RPEr           &\underline{0.378}&\underline{0.700}&\textbf{0.151}&\textbf{0.06}   &\textbf{0.067}  &\textbf{0.094}  
                                     &\textbf{0.065} &\underline{0.298}&\underline{0.343}&\textbf{0.167}           
                                     &\textbf{0.232} \\
                    & ATE            &\textbf{0.106}  &\textbf{0.074}  &\textbf{0.082}  &\textbf{0.085}  &\textbf{0.081} &\textbf{0.092}  
                                     &\textbf{0.109}  &\textbf{0.108}  &\textbf{0.102}  &\textbf{0.108}          
                                     &\textbf{0.095} \\

                \bottomrule      
            \end{tabular}}}
	\end{center}
    \begin{tablenotes}    
        \footnotesize               
        \item[1] All displayed data of SLAM and VIO systems except for SVO2.0 are derived from the median of ten runs.           
        \item[2] Average error of the successful sequences. Systems that did not complete all sequences are marked with *.      
        \item[3] The data of SVO2.0 are referenced from~\cite{wang2022lf}.
    \end{tablenotes}
    \end{threeparttable}
\end{table*}

\begin{figure*}[h]
    \centering
    
    \subfigure{
        \begin{minipage}[t]{0.9\linewidth}
            \centering
            \includegraphics[width=1\linewidth]{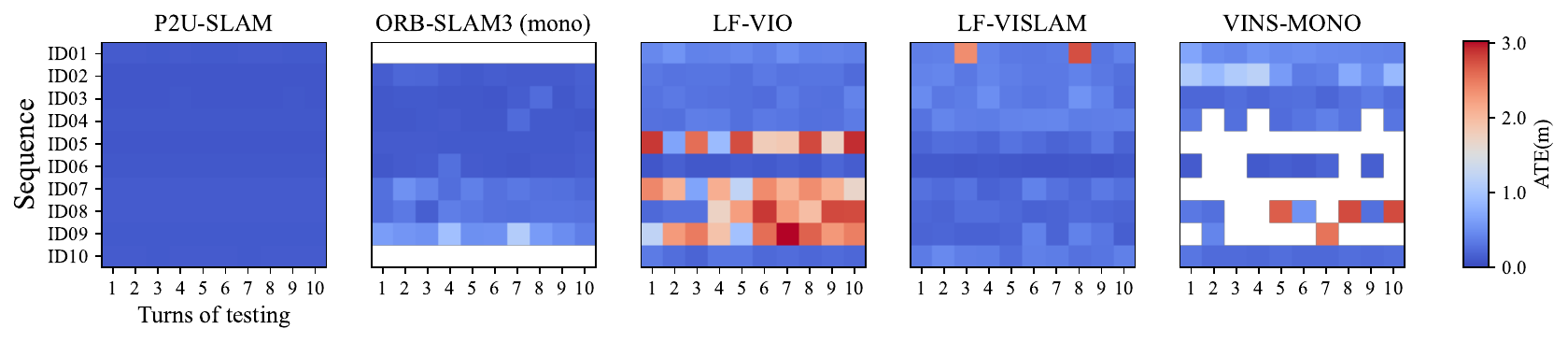}
        \end{minipage}
    }
    \vspace{-4mm}
    
    \subfigure{
        \begin{minipage}[t]{0.9\linewidth}
            \centering
            \includegraphics[width=1\linewidth]{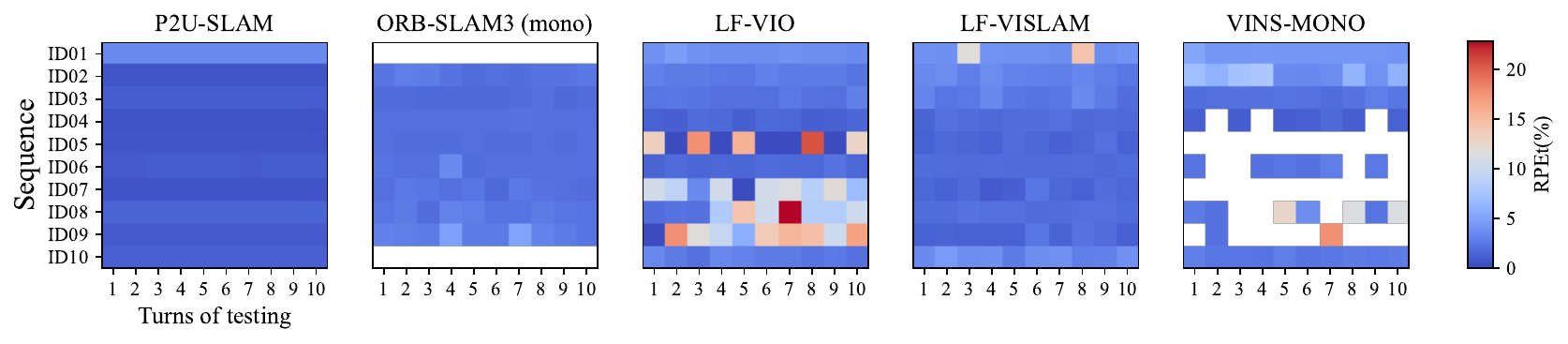}
        \end{minipage}
    }
    \vspace{-4mm}
    
    \subfigure{
        \begin{minipage}[t]{0.9\linewidth}
            \centering
            \includegraphics[width=1\linewidth]{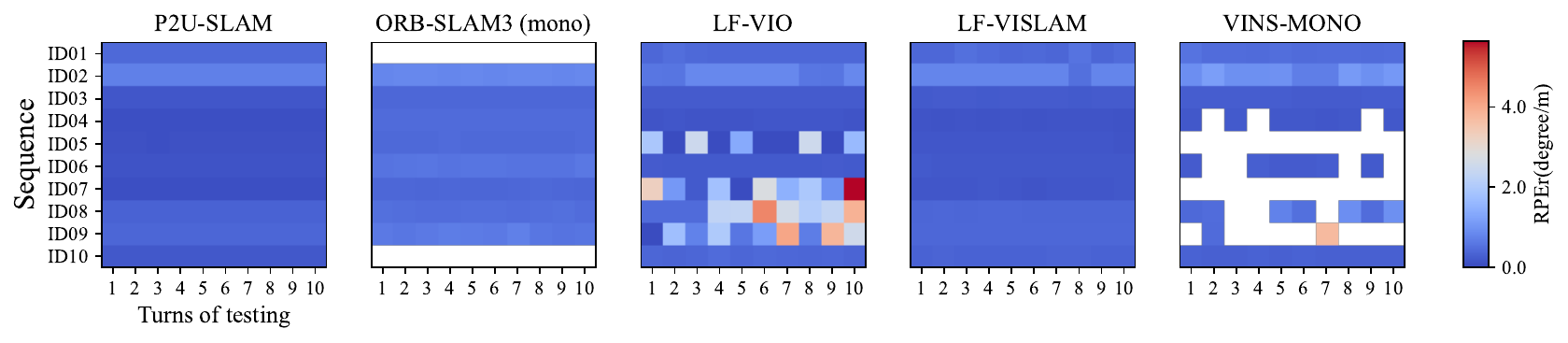}
        \end{minipage}
    }
    \vspace{-2mm}
    \caption{The robustness of several algorithms on the PALVIO dataset is demonstrated, mainly through three metrics: a) RMS ATE (m), b) RMS RPEt ($\%$), and c) RMS RPEr (degree/m). Each algorithm runs a total of ten times on each sequence, with the results of each run represented by the color squares. A higher \CSRED{red} component in the square indicates a worse result in the corresponding metric test, while a greater \CSBLUE{blue} component signifies better performance. An incomplete run is represented by a blank white square.
    \add{It should be noted that the incomplete runs of ORB-SLAM3 (mono) are slightly different from those of VINS-MONO. In the monocular mode, ORB-SLAM3 has no IMU input. Once the visual constraints fail, the loss of positioning cannot be recovered, and there is no difference in the results of repeated experiments on the same failure sequence. However, VINS-MONO still has the motion constraints of the IMU when the visual constraints fail. Therefore, there will be inconsistent results in repeated experiments on the same sequence.}}
    \label{fig:palvio_dataset_robustness}
\end{figure*}

\subsection{Evaluation on PALVIO Dataset}
The PALVIO dataset consists of $10$ indoor sequences, recorded using two hand-held PAL cameras (FoV: $360^{\circ}{\times}(40^{\circ}{\sim}120^{\circ}$, resolution: $1280{\times}960$)) and an IMU sensor ($200Hz$). 
With the data captured by the motion capture system as the ground truth, 
we compare our P2U-SLAM system with four open-source state-of-the-art vSLAM/VIO methods SVO2.0~\cite{forster2016svo}, VINS-Mono~\cite{qin2018vins}, ORB-SLAM3 (monocular mode)~\cite{campos2021orb}, LF-VIO~\cite{wang2022lf}, and LF-VISLAM~\cite{wang2023lfvislam}. 
Absolute Translation Error (ATE) and Relative Pose Error (RPE) are used as the evaluation metrics for assessment. 

As shown in Table~\ref{EVAL_on_PALVIO_DATASET}, P2U-SLAM has the lowest average error in ATE and RPE metrics. 
In detail, P2U-SLAM reaches the best precision on all sequences in ATE and RPEt metrics, and most sequences in RPEr metrics. 
On a handful of sequences where P2U-SLAM ranks 2nd in RPEr metrics, the SLAM/VIO methods (SVO2.0, LF-VIO, and LF-VISLAM) performing best are all fusing IMU measurements, while P2U-SLAM reaches comparable precision with the only monocular visual PAL input. 
Furthermore, ID02, ID04, and ID08 are chosen as representative sequences for exhibition in Fig.~\ref{fig:palvio_dataset_trans_rot_error}.

Apart from SVO2.0 whose data are cited from~\cite{wang2022lf}, all other SLAM/VIO frameworks are tested $10$ times on the PALVIO dataset and the outcome of experiments is expressed as colored squares in Fig.~\ref{fig:palvio_dataset_robustness} to evaluate the robustness of these frameworks. 
The fewer turns of failure there are, the more robust the corresponding system is. 
P2U-SLAM, LF-VIO, and LF-VISLAM are naturally able to process map points from the negative half-plane to fully utilize the whole FoV of the PAL image, due to the adaptation to the Taylor camera model. 
The larger FoV has made a great contribution to stably tracking in rapid rotation and motion. 
Therefore, P2U-SLAM, LF-VIO, and LF-VISLAM succeed in every trial of experiments. 
However, the robustness performance of these three methods is not the same.
With the help of point and pose uncertainty, the P2U-SLAM can suppress the influence on optimization from outliers or weak constraints,
and P2U-SLAM finally achieves the lowest variance in ten times of testing. 

In summary, P2U-SLAM has been verified to be robust to wide-FoV PAL input and notably stands out among the compared methods in this dataset with all methods under Sim3 alignment.

\begin{table}[h] 
	\caption{Performance comparison on the public TUM-VI benchmark~\cite{schubert2018tum} (RMS ATE in $m$).}        
	\label{EVAL_on_TUMVI}           
    \begin{threeparttable}
	\begin{center}                 
            \renewcommand\arraystretch{1.1}{\setlength{\tabcolsep}{2.0mm}{\begin{tabular}{c | c c c c c | c}
			\toprule               
                \multirow{3}{*}{Sequences} 
                                     &VINS-           &\add{LF-}       &LF-             &ORB-           &P2U-             & Length\\
                                     &Mono            &\add{VISLAM}    &VIO             &SLAM3$^1$      &SLAM             & (m)\\
                                     &\cite{qin2018vins} &\cite{wang2023lfvislam} &\cite{wang2022lf} &\cite{campos2021orb} &(ours)\\

                \midrule              
			  room1                &0.089           &\add{0.089}     &0.094           &0.074           &\textbf{0.071}  &146  \\
			  room2                &0.107           &\add{0.117}     &0.129           &0.048           &\textbf{0.047}  &142  \\
                room3                &0.129           &\add{0.123}     &0.114           &0.060           &\textbf{0.049}  &135  \\
			  room4                &0.251           &\add{\textbf{0.059}}  &0.361     &0.078           &0.079           &68   \\
                room5                &0.219           &\add{0.174}     &0.220           &\textbf{0.072}  &0.075           &131  \\
                room6                &0.097           &\add{0.113}     &0.120           &\textbf{0.073}  &0.074           &67   \\
                \midrule  
                corridor1            &0.771           &\add{0.476}     &0.380           &0.358           &\textbf{0.065}  &305  \\
			  corridor2            &0.872           &\add{0.569}     &0.413           &0.073           &\textbf{0.065}  &322  \\
                corridor3            &0.635           &\add{0.638}     &0.585           &0.269           &\textbf{0.070}  &300  \\
			  corridor4            &0.186           &\add{0.520}     &0.471           &0.231           &\textbf{0.069}  &114  \\
                corridor5            &0.845           &\add{0.487}     &0.484           &0.085           &\textbf{0.060}  &270  \\
                \midrule  
                magistrale1          &18.68           &\add{0.747}     &4.805           &1.933           &\textbf{0.085}  &918  \\
			  magistrale2          &1.956           &\add{0.531}     &0.543           &0.748           &\textbf{0.086}  &561  \\
                magistrale3          &0.756           &\add{0.534}     &1.894           &0.969           &\textbf{0.086}  &566  \\
			  magistrale4          &1.025           &\add{0.736}     &0.780           &0.502           &\textbf{0.084}  &688  \\
                magistrale5          &1.690           &\add{0.885}     &1.135           &1.331           &\textbf{0.081}  &458  \\
                magistrale6          &4.032           &\add{0.727}     &4.167           &0.176           &\textbf{0.093}  &711  \\
                \bottomrule      
            \end{tabular}}}
	\end{center}
    \begin{tablenotes}    
        \footnotesize               
        \item[1] ORB-SLAM3 is set in the monocular mode.
    \end{tablenotes}
    \end{threeparttable}
\end{table}
\subsection{Evaluation on TUM-VI Benchmark}
Three types of sequences of room, corridor, and magistrale are selected from the TUM-VI dataset to evaluate the performance of P2U-SLAM in various complicated environments such as rapid motion, duplicate textures, scene switches, and brightness changes. 
The resolution and FoV of the fisheye image input are respectively set $512{\times}512$ and $360^{\circ}{\times}(0^{\circ}{\sim}95^{\circ})$, and the frequency of IMU data is $200Hz$ on the TUM-VI benchmark. 
The ground truth captured by the motion capture system is available for complete trajectory on room sequences, while unavailable except at the starting and ending points on corridor and magistrale sequences. 
Therefore, ATE serves as a suitable metric to measure the cumulative drift along the whole trajectory.

\begin{figure}[t]
    \centering
    \includegraphics[width=3.4in]{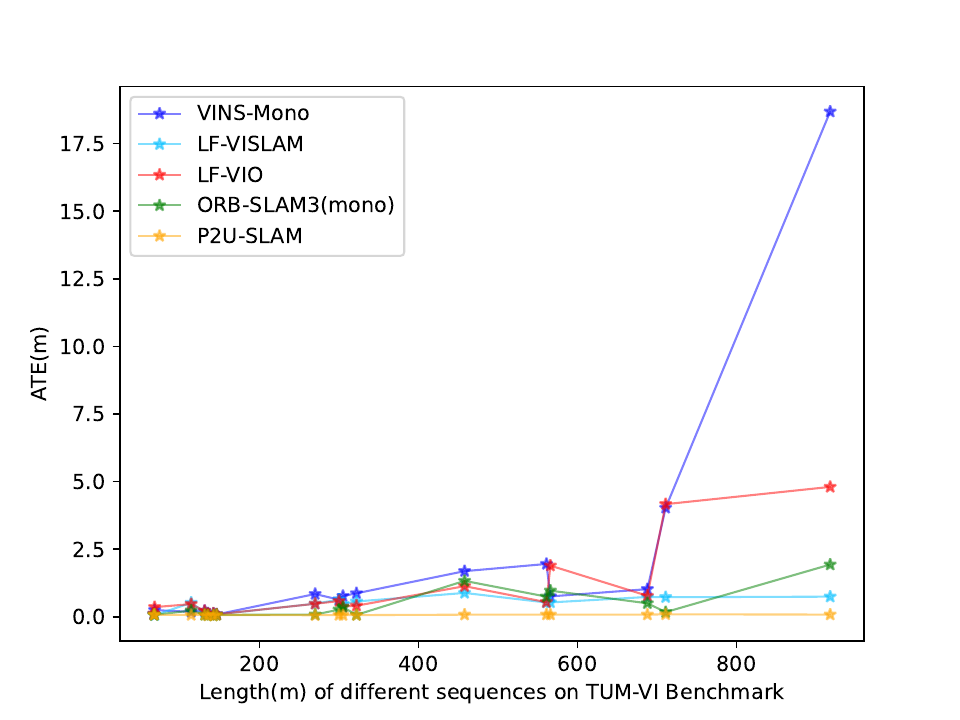}
    \caption{Stability comparison on sequences from TUM-VI dataset. Each node on the line chart corresponds to a sequence in the dataset.}
    \label{fig_tum_robust}
    \end{figure}

The RMS ATE results obtained are compared with the most relevant open-source systems in Table~\ref{EVAL_on_TUMVI}. 
On room sequences, P2U-SLAM achieves comparable precision which is less than $0.1m$ with the state-of-the-art framework ORB-SLAM3 (monocular)~\cite{campos2021orb}. 
However, the cumulative drift of the ORB-SLAM3 increases significantly with the length of the test sequence on corridor and magistrale sequences like other VIO methods, while the P2U-SLAM always maintains the ATE of no more than $0.1m$ on every sequence. 
Considering that P2U-SLAM and ORB-SLAM3 use the same feature extraction method and multi-thread framework, the results in Fig.~\ref{fig_tum_robust} illustrate that the uncertainty method adopted by P2U-SLAM achieves a high degree of error stability in trajectory tracking within one kilometer.
\add{It is also worth noting that, with the support of the complete loop closure function and IMU data, LF-VISLAM also maintains excellent consistency on the TUM-VI sequences with multi-scale changes. However, it is still slightly inferior compared to P2U-SLAM, which has the support of dual uncertainties.}

\begin{figure}[h]
    \centering
    \includegraphics[width=3.4in]{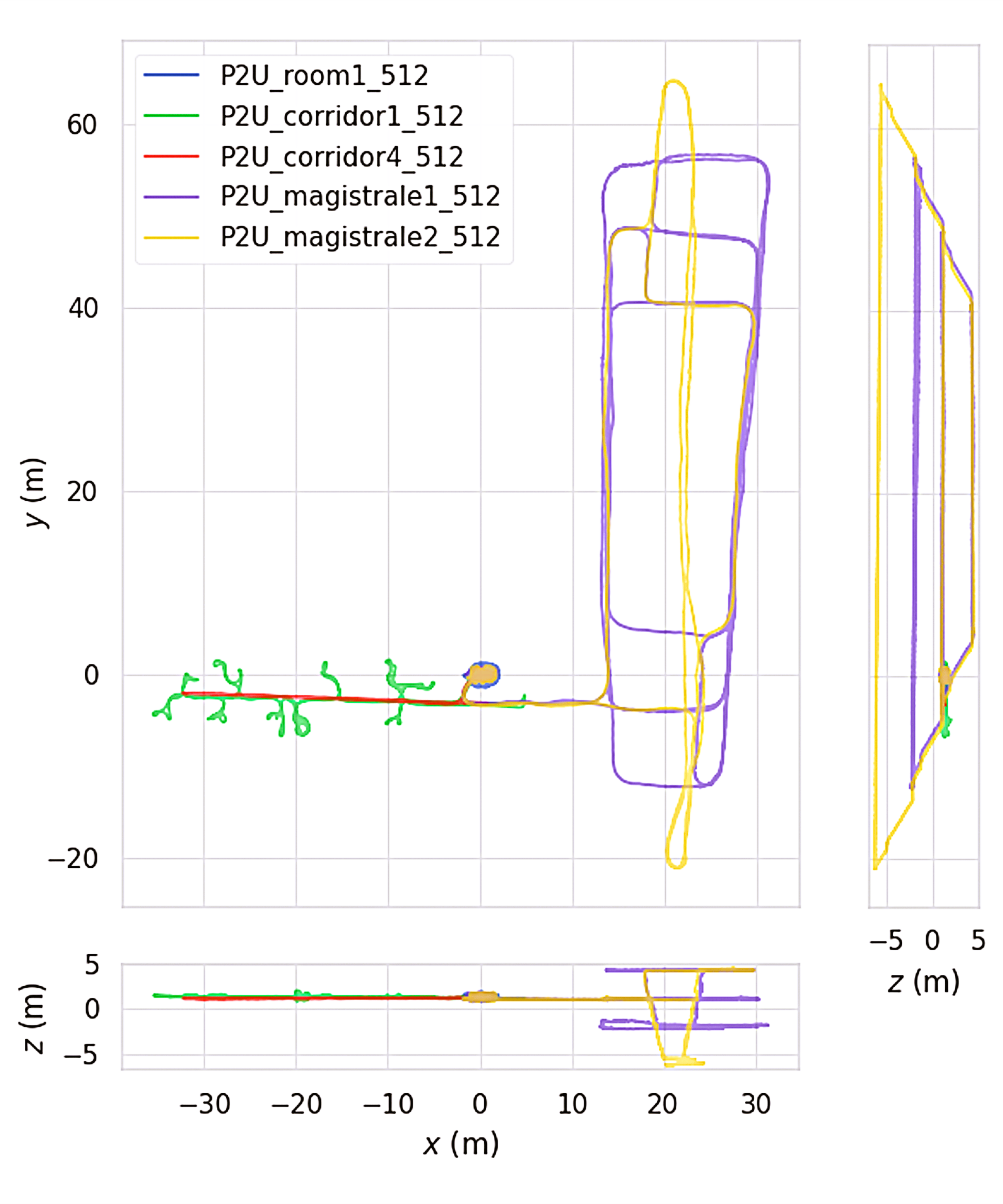}
    \caption{Multi-session P2U-SLAM results with several sequences from the TUM-VI dataset (top, side, and front views).}
    \label{fig_tum_3iew}
\end{figure}

Part of the multi-session experimental results on the TUM-VI dataset for P2U-SLAM are displayed in Fig.~\ref{fig_tum_3iew}. 
This figure is instrumental in demonstrating the system's capabilities in handling real-world scenarios with varying complexities. 
The top, side, and front views presented in the figure offer a three-dimensional perspective of the trajectory followed by the system, showcasing the accuracy and robustness of the P2U-SLAM in navigating through different environments. 
The figure highlights the system's ability to maintain consistency in trajectory mapping even in challenging conditions, which is a testament to the effectiveness of the point and pose uncertainty integration within the SLAM framework.
Additionally, the experiment also demonstrates the map reuse and expansion capabilities of P2U-SLAM.

\subsection{Ablation Study}
To further evaluate the impact of point and pose uncertainty on the performance of P2U-SLAM, we conducted an ablation study on sequences corridor1 to corridor5 from the TUM-VI benchmark. 
As depicted in Table~\ref{abaltion_study}, the outcome of the ablation study is presented with the $5$ dimensions of ATE for different configurations: without any uncertainty, only with point uncertainty, only with pose uncertainty, and with both enabled. 
It should be noted that the extreme values of ATE at a particular moment are susceptible to the real-time performance of the operating computer or the impact of thread blocking. 
It is acceptable that the configuration of applying both point and pose uncertainty does not yield the smallest error in the Min ATE metric. 
More importantly, in the evaluation of the other four dimensions of ATE, not only is the configuration that applies a single uncertainty better than not applying it, but the results of applying both types of uncertainty are also better than those obtained by applying only one type of uncertainty.

\begin{table}[!t] 
	\caption{Ablation studies for point and pose uncertainty with metric ATE ($m$) on sequences corridor1$\sim$5 of TUM-VI benchmark~\cite{schubert2018tum}.}        
	\label{abaltion_study}           
	\begin{center}                 
            \renewcommand\arraystretch{1.1}{\setlength{\tabcolsep}{2.8mm}{\begin{tabular}{c c | c c c c c}
			\toprule               
                ${\Sigma}_{ip}^{\prime}$  &${\Sigma}_{pj}^{\prime\prime}$    &RMS & Mean & Median & Min &Max\\

                \midrule  
                                &              &0.0705    &0.0652   &0.0628 &0.0366          &0.266        \\
			\checkmark      &              &0.0655    &0.0609   &0.0587 &0.0178          &0.214        \\
			                &\checkmark    &0.0659    &0.0614   &0.0585 &\textbf{0.0166} &0.204     \\
                \cellcolor{gray!20}\checkmark  &\cellcolor{gray!20}\checkmark  
                &\cellcolor{gray!20}\textbf{0.0642} &\cellcolor{gray!20}\textbf{0.0597} &\cellcolor{gray!20}\textbf{0.0573} &\cellcolor{gray!20}0.0178  
                &\cellcolor{gray!20}\textbf{0.202}\\
                \bottomrule      
            \end{tabular}}}
	\end{center}
    \begin{tablenotes}    
        \footnotesize               
        \item[] ${\Sigma}_{ip}^{\prime}$ and ${\Sigma}_{pj}^{\prime\prime}$ respectively stand for point and pose uncertainty as shown in Eq.~\eqref{Covariance_Point} and Eq.~\eqref{Covariance_POSE}. .
    \end{tablenotes}
\end{table}

\subsection{Computing Time Analysis}
In this subsection, we report the efficiency of P2U-SLAM. 
Table~\ref{Computing_Time} compares the running time of the main operations performed in P2U-SLAM with ORB-SLAM3, indicating that P2U-SLAM owns the ability to work in real-time at about $40{\sim}60$ frames and $3{\sim}6$ keyframes per second. 
Nevertheless, the average running time of P2U-SLAM is slightly larger than ORB-SLAM3 (mono) with around $22\%$ and $14\%$ in respectively the tracking and local mapping module, since the introduction of the uncertainty module in optimization of P2U-SLAM.
This efficient performance highlights P2U-SLAM's capability to maintain high accuracy with comparable speed, making it ideal for real-time applications.

\begin{table}[h] 
	\caption{Running time of the main parts of P2U-SLAM compared to ORB-SLAM3 (mono)~\cite{campos2021orb}, measured on TUM-VI corridor2~\cite{schubert2018tum} (mean time and standard deviation in $ms$).}        
	\label{Computing_Time}           
	\begin{center}                 
            \renewcommand\arraystretch{1.1}{\setlength{\tabcolsep}{2.8mm}{\begin{tabular}{c | c c }
			\toprule               
                Module                &P2U-SLAM  &ORB-SLAM3 ~\cite{campos2021orb}\\

                \midrule              
			Tracking                &20.68$\pm$7.84    &16.97$\pm$4.77               \\
			Local mapping           &198.13$\pm$129.87 &174.36$\pm$102.66            \\
                Place recognition       &5.25$\pm$2.83     &6.12$\pm$4.08                \\
			Loop Fusion             &42.82$\pm$25.56   &43.61$\pm$27.20              \\
                Loop Global BA          &451.06$\pm$189.24 &447.25$\pm$186.82            \\
                \midrule 
                Uncertainty             &\multirow{2}{*}{0.83$\pm$0.55} &\multirow{2}{*}{/}             \\
                Calculation \& Updating   & &             \\
                \bottomrule      
            \end{tabular}}}
	\end{center}
\end{table}

\section{Conclusion}
\label{sec:conclusion}

In this study, we propose P2U-SLAM, a framework for wide-FoV cameras that makes use of point and pose uncertainty to overcome weak feature correspondence introduced by a wide overlap area. 
The proposed system establishes a complete visual SLAM framework, which includes initialization, tracking, local mapping, loop closing, and map merging modules. 
To handle 3D points from the negative place ($z{<}0$), the system inherits the Taylor camera model used in previous work. 
In addition, this work shows that it is mathematically justified and necessary to incorporate point and pose uncertainty during optimization. 
Moreover, the calculation and use of point and pose uncertainty are integrated into the SLAM framework, granting P2U-SLAM high positioning accuracy and robustness that exceeds the state-of-the-art methods on two public wide-FoV SLAM datasets.
There is still a lot of room for improvement in P2U-SLAM. 
Despite a significant decrease in cumulative drift in the kilometer-level series, P2U-SLAM is still a pure visual SLAM framework. 
It is challenging for P2U-SLAM to work in some scenarios such as elevators and subways. 
Besides, the visualization of the map generated by P2U-SLAM is sparse and limited in downstream applications compared to dense maps.

In the future, we intend to extend P2U-SLAM to the visual-inertial framework to enhance the robustness of our algorithm in more extreme scenarios where vision fails. 
In addition, the uncertainty method of P2U-SLAM is strongly related to the current active research on 3D Gaussian Splatting, so achieving real-time dense mapping with higher efficiency from P2U-SLAM is also one of our exploration directions.

\bibliographystyle{IEEEtran}
\bibliography{reference}

\end{document}